\documentclass{article}
\usepackage{iclr2027_conference,times}

\usepackage{array}
\usepackage{graphicx}
\usepackage{wrapfig}
\usepackage{amsmath,amssymb}
\usepackage{algorithm,algpseudocode}
\usepackage{xspace}
\usepackage{multirow}
\usepackage{capt-of}
\usepackage{booktabs}
\usepackage{etoolbox}
\usepackage[table]{xcolor}

\definecolor{ourscolor}{HTML}{CFE6DB}

\newcommand{\figref}[1]{Fig.~\ref{#1}}
\newcommand{\tabref}[1]{Tab.~\ref{#1}}
\newcommand{\secref}[1]{Sec.~\ref{#1}}
\newcommand\mypara[1]{\noindent\textbf{#1}}
\newcommand{\E}{\mathbb{E}}

\newcolumntype{x}[1]{>{\centering\arraybackslash}p{#1}}
\newcolumntype{y}[1]{>{\raggedright\arraybackslash}p{#1}}
\let\originalleft\left
\let\originalright\right
\renewcommand{\left}{\mathopen{}\mathclose\bgroup\originalleft}
\renewcommand{\right}{\aftergroup\egroup\originalright}

\usepackage[hidelinks]{hyperref}
\hypersetup{pdftitle={UniLAP: A Unified Latent Action Policy for Cross-Embodiment Manipulation},pdfauthor={Lei Yao, Qiao Feng, Mutian Tong, Qihao Yun, Lap-Pui Chau, Lingjie Liu}}

\let\cite\citep

\makeatletter
\patchcmd{\@outputpage}{\box\@tempboxa}{\pdfrunninglinkoff\box\@tempboxa\pdfrunninglinkon}{}{\PackageError{unilap}{Header link patch failed}{}}
\patchcmd{\@outputpage}{\baselineskip\footskip}{\baselineskip\footskip\pdfrunninglinkoff}{}{\PackageError{unilap}{Footer link patch failed}{}}
\AddToHook{cmd/footnoterule/before}{\pdfrunninglinkoff}
\makeatother

\newcommand{\ours}{UniLAP\xspace}

\title{\ours: A Unified Latent Action Policy \\ for Cross-Embodiment Manipulation}

\newcommand{\institutionskip}{\quad}
\author{Lei Yao\textsuperscript{1,2}\textsuperscript{\dag}, Qiao Feng\textsuperscript{1}, Mutian Tong\textsuperscript{1}, Qihao Yun\textsuperscript{1}, Lap-Pui Chau\textsuperscript{2}, Lingjie Liu\textsuperscript{1} \\
  \textsuperscript{1}University of Pennsylvania \institutionskip
  \textsuperscript{2}The Hong Kong Polytechnic University \\
  {\tt\small \url{https://rayyoh.github.io/UniLAP/}}
}

\iclrfinalcopy

\begin{document}
\maketitle
\fancyhead{}
\fancyhead[L]{Preprint}
\raggedbottom
\renewcommand{\floatpagefraction}{0.8}
\setcounter{topnumber}{4}
\setcounter{bottomnumber}{2}
\setcounter{totalnumber}{6}
\setlength{\textfloatsep}{14pt plus 2pt minus 2pt}
\setlength{\floatsep}{10pt plus 2pt minus 2pt}
\setlength{\intextsep}{8pt plus 2pt minus 2pt}
\begingroup
\renewcommand{\thefootnote}{\fnsymbol{footnote}}
\footnotetext[2]{Work done during a visit to the University of Pennsylvania.}
\endgroup

\begin{abstract}
    Latent Action Models (LAMs) can capture manipulation priors from action-free videos, but discrete latents usually impose a granularity bottleneck, limiting their ability to represent fine-grained action. Recent continuous alternatives mitigate this, but they are typically used as proxies for world modeling and require a further planning step to produce executable actions. We thus introduce \ours{}, a unified continuous latent-action policy that models the latent space via flow matching and bypasses reliance on future frames during inference. By distilling a LAM teacher with a flow-adaptive objective, \ours{} inherits the teacher's action representation while retaining the multimodal distribution of feasible behaviors. To decouple latent representations from task-irrelevant appearance, we extract a label-free motion target to isolate relevant kinematic cues from spurious visual dynamics. With the established embodiment-agnostic latent manifold, \ours{} can be easily adapted to novel platforms. Pretrained on multi-source videos at a compact 1.1B parameter scale, \ours{} outperforms state-of-the-art latent-action and Vision-Language-Action (VLA) baselines across hand pose forecasting, single-arm and humanoid manipulation, and real-world deployment.
\end{abstract}

\begin{figure}[!h]
    \centering
    \includegraphics[width=0.9\linewidth]{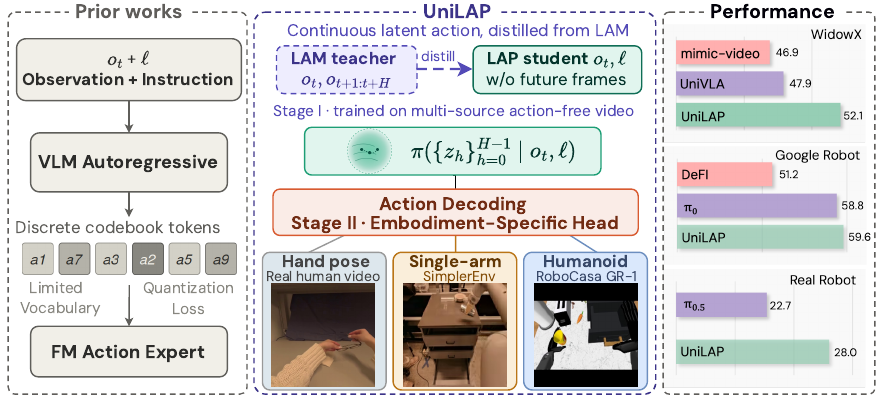}
    \caption{\textbf{\ours{} learns a continuous unified latent action policy (LAP) from action-free video.} \emph{Left}, prior pipelines quantize the latent into discrete codebooks. \emph{Middle}, \ours{} models it as a continuous flow distilled from a frozen LAM teacher. \emph{Right}, performance comparison.}
    \label{fig:teaser}
\end{figure}

\section{Introduction}
\label{sec:intro}
Action-labeled data is expensive to collect at scale, and this scarcity limits the development of generalist manipulation policies~\cite{zheng2025survey}. Action-free human and robot videos, by contrast, are abundant and dense with motion and dynamic cues. Latent Action Models (LAMs)~\cite{schmidt2023learning,bruce2024genie,ye2024latent} exploit such videos via extracting latent actions from consecutive frames by next-frame prediction without requiring explicit action annotations. Within Vision-Language-Action (VLA) systems, an action expert is then learned on limited labeled data to decode these latents into specific robot commands~\cite{bu2025univla,chen2025moto}, separating the data-hungry acquisition of action priors from their data-scarce grounding on a target robot.

Despite their potential, existing LAM pipelines~\cite{ye2024latent,bu2025univla,chen2025moto,yang2026chain} suffer from two key limitations. First, most methods quantize them into discrete tokens and predict them via large VLMs autoregressively (\figref{fig:teaser} \textit{left}). This discretization coarsens the latent onto the nearest code, hindering its ability to express the continuous, multimodal distribution that manipulation requires. Recent works~\cite{gao2025adaworld,gao2026dreamdojo} alternatively loosen such representational bottleneck and illustrate the effectiveness of continuous latent actions. However, these latents are usually inferred with access to future frames and are deployed by world-model prediction followed by planning~\cite{goswami2025world,assran2025v}, leaving open how to turn continuous latent actions into a causal policy for robot control. Second, the standard pixel reconstruction latent learning entangles it with task-irrelevant appearance, a problem amplified by frequent camera motion. Prior work attempts to decouple this by factorizing the latent into separate tokens~\cite{bu2025univla,li2025latbot} or introducing optical flow~\cite{bu2025laof,bi2025motus}, but the former requires a dedicated design to enforce the separation, while the latter relies on an external optical flow estimator to generate dense motion targets for latent action learning.

We present \ours{}, a unified latent action policy that addresses these limitations. \textbf{(1)~Continuous latent action policy.} Such a representation preserves the multimodal distribution that a discrete codebook cannot describe. We model the latent action as a continuous flow~\cite{lipman2022flow} conditioned on current observation, and distill the LAM teacher's posterior via a \emph{flow-adaptive objective}. This mechanism tightens the latent toward the teacher as the noise decreases. Acting without future frames, our method shortens the route from action-free pretraining to robot control. \textbf{(2)~Label-free motion supervision.} Pixel reconstruction supervision binds the latent to appearance, making camera movement contaminate it. \ours{} instead derives a label-free motion target from cross-frame correspondences in a frozen ViT, isolating task-relevant information from unrelated visual changes. This supervision needs neither explicit factorization nor an external flow estimator. \textbf{(3)~Cross-embodiment adaptation.} The continuous latent forms an embodiment-agnostic action space shared across human and multi-robot videos. A new robot reuses this pretrained latent and adapts with only a fresh action decoder and light policy finetuning, rather than per-robot retraining.

We evaluate \ours{} on hand pose forecasting~\cite{hoque2025egodex}, single-arm manipulation~\cite{li2024evaluating}, humanoid manipulation~\cite{bjorck2025gr00t}, and tabletop manipulation on a real robot platform. At a smaller $1.1$B scale, \ours{} compares favorably with latent-action and VLA baselines, surpassing recent action-free pretraining methods. Analyses and ablations further support the importance of continuous latent-action modeling, flow-adaptive posterior distillation, and correspondence-based motion supervision for learning transferable action representations.

\section{Related Work}
\label{sec:related}

\mypara{Policy Learning from Videos.}
Videos inherently capture rich dynamic information about the world transitions, making them valuable for robot learning. Early contributions have provided effective visual representations from videos via self-supervision, which can be transferred to policy learning. Another line of work incorporates world models~\cite{ha2018world}, aiming to estimate future states and derive action via motion planning~\cite{goswami2025world,assran2025v} or inverse dynamic models~\cite{kim2025uniskill,chen2025large}. Instead, recent world action models~\cite{liang2025video,hu2024video,pai2025mimic,ye2026world} are proposed to co-train video generation with action prediction, treating video latents as structured variables for underlying dynamics. Meanwhile, other works like EgoMimic~\cite{kareer2025egomimic} and EgoScale~\cite{zheng2026egoscale} explore bridging the domain gap between egocentric videos and robot execution, treating humans as a special embodiment and hand motion as an action proxy. Our work differs from these in that we model a unified latent policy aligning various video sources without requiring explicit action supervision, and then transfer the latent action across embodiments.

\mypara{Latent Action Learning.}
Latent Action Models (LAM) are normally learned from consecutive frames without explicit action annotations~\cite{schmidt2023learning,bruce2024genie}. The dominant paradigm~\cite{ye2024latent,bu2025univla,chen2025moto,yang2026chain} casts latent actions as \emph{intermediate representation} within vision-language-action (VLA) frameworks, with flow-matching~\cite{lipman2022flow,black2024pi_0} or diffusion~\cite{chi2025diffusion,yin2025diverse} action experts. However, discrete latent actions via VQ-VAE~\cite{van2017neural} limit action granularity to the codebook size. IGOR~\cite{chen2024igor} and CLAM~\cite{liang2025clam} learn continuous latents via variational objectives, while AdaWorld~\cite{gao2025adaworld} adopts a $\beta$-VAE bottleneck that DreamDojo~\cite{gao2026dreamdojo} scales to large-scale data. To address motion-scene entanglement, LatBot~\cite{li2025latbot} factorizes the latent into separate tokens, while UniVLA~\cite{bu2025univla} learns to filter out irrelevant information. Villa-X~\cite{chen2025villa} couples latent and real action diffusion with two experts, while others~\cite{jiang2026cross,zhang2025align} directly encode real actions across embodiments. We instead transplant the generative formulation onto \emph{latent actions}, yielding a shared \emph{latent flow-matching policy} across embodiments.

\mypara{Visual Foresight for Manipulation.}
Instead of just relying on static image-action pairs, visual foresight has been shown to boost policy learning by implicitly encoding necessary dynamics. FLARE~\cite{zheng2025flare} aligns learnable future tokens with vision-language embeddings of a subsequent observation, while LDA-1B~\cite{lyu2026lda} co-denoises action and upcoming visual latents. DreamVLA~\cite{zhang2025dreamvla} broadens the signal across modalities, forecasting depth and semantics simultaneously. LAOF~\cite{bu2025laof} augments pixel reconstruction with optical flow to disentangle physical motion from photometric change, while Motus~\cite{bi2025motus} entirely uses optical flow targets. These methods rely on external flow estimators and thus inherit their biases. In contrast, \ours{} derives a dense \emph{correspondence residual} intrinsically from key vectors in a frozen ViT, requiring no auxiliary network to generate targets for latent policy pretraining.

\section{Methodology}
\label{sec:method}
\begin{figure*}
    \centering
    \includegraphics[width=0.99\linewidth]{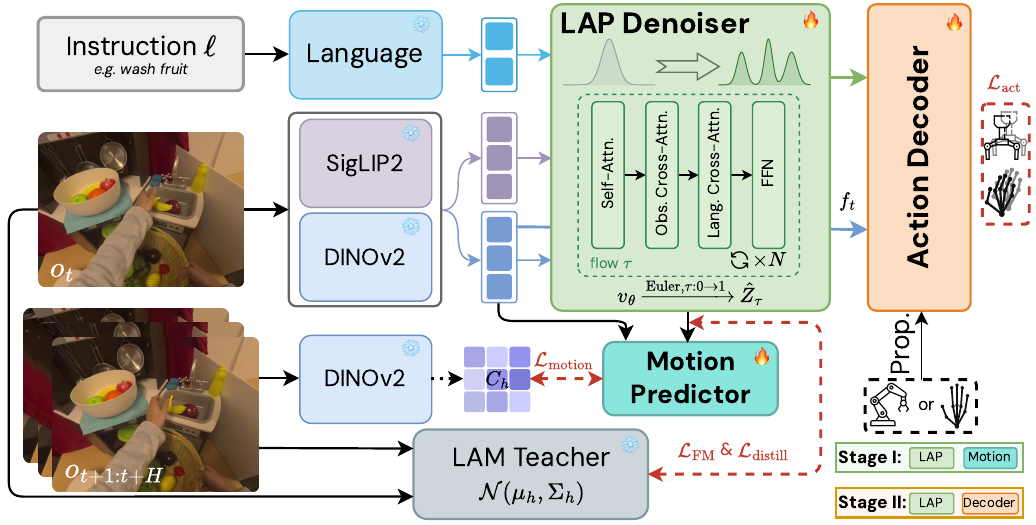}
    \caption{\textbf{Overview of \ours.} \emph{Stage~I} pretrains a continuous flow-matching latent action policy (LAP) on action-free videos under $\mathcal{L}_{\text{FM}}$. A frozen LAM teacher's posterior is distilled via the flow-adaptive $\mathcal{L}_{\text{distill}}$, and a cross-frame correspondence target supervises the motion predictor to regularize the latent. \emph{Stage~II} attaches a per-embodiment decoder $\phi_{\text{dec}}$, trains it from scratch, and finetunes the latent policy on the target embodiment's labeled actions.}
    \label{fig:overview}
\end{figure*}

In this paper, we present \ours{}, which learns a shared latent policy from action-free videos across embodiments. It comprises three components trained in two stages. In Stage~I, the \emph{unified latent action policy} (\secref{sec:lap}) models latent actions as a continuous flow-matching process distilled from a frozen LAM teacher, while the \emph{correspondence-driven motion supervision} (\secref{sec:cms}) regularizes the latent space with a label-free target. In Stage~II, an \emph{action decoder} (\secref{sec:lad}) maps the latents into executable commands through an embodiment-specific head. \figref{fig:overview} shows the overall architecture.

\subsection{Unified Latent Action Policy}
\label{sec:lap}
Prevailing methods~\cite{bu2025univla} predict discrete latent action tokens, fixing granularity at the codebook size, while the continuous latents of recent LAMs~\cite{gao2026dreamdojo} are more expressive but inferred with access to future frames, hence not directly usable as a causal policy.

At time $t$, the policy predicts a horizon-$H$ latent action chunk from observation $o_t$ and instruction $\ell$, while the LAM teacher reads the frame window $\{o_{t+j}\}_{j=0}^{H}$. Let $\tau \sim \mathcal{U}(0,1)$ be the flow timestep, $h \in \{0, \dots, H{-}1\}$ the chunk index, and $x_{0,h} \sim \mathcal{N}(0,I)$ the initial noise. The latent action $z_h \in \mathbb{R}^{d_z}$ encodes the transition $(o_{t+h}, o_{t+h+1})$, whose non-causal Gaussian posterior from LAM is:
\begin{equation}
    q(z_h \mid o_{t+h}, o_{t+h+1}) = \mathcal{N}(\mu_h, \Sigma_h),
\end{equation}
where $\mu_h$ and diagonal covariance $\Sigma_h$ come from the LAM encoder on $(o_{t+h}, o_{t+h+1})$.
The latent policy $\pi(\{z_h\}_{h=0}^{H-1} | o_t, \ell)$ outputs the entire $H$-step chunk in a single forward pass, though we keep per-step notation for brevity. We parameterize it as a flow-matching network~\cite{lipman2022flow}, following the paradigm in visuomotor policies~\cite{chi2025diffusion,black2024pi_0}. Two frozen ViTs encode $o_t$, DINOv2~\cite{oquab2023dinov2} for fine-grained spatial features and SigLIP2~\cite{tschannen2025siglip} for vision-language semantics, and a pre-trained language backbone embeds $\ell$. A stack of transformer blocks conditions on the flow timestep $\tau$ through Adaptive Layer Normalization (AdaLN)~\cite{perez2018film,peebles2023scalable} and visual and language tokens via per-block cross-attention, predicting the velocity field $v_\theta$ along the linear interpolant $x_{\tau,h} = (1-\tau)x_{0,h} + \tau z_h$. With target $z_h \sim q(\cdot \mid o_{t+h}, o_{t+h+1})$, we optimize the policy under the flow-matching objective:
\begin{equation}
    \mathcal{L}_{\text{FM}} = \mathbb{E}_{\tau, x_{0,h}, z_h} \!\left[ \bigl\| v_\theta(x_{\tau,h}, o_t, \ell) - (z_h - x_{0,h}) \bigr\|_2^2 \right].
\end{equation}
The policy is conditioned only on $(o_t, \ell)$, whereas the teacher infers latent actions using future frames. $\mathcal{L}_{\text{FM}}$ therefore targets the teacher posterior marginalized over these unobserved frames.
\begin{equation}
    p^{*} \left(\{z_h\}_{h=0}^{H-1} \mid o_t, \ell \right) = \E_{\{o_{t+j}\}_{j=1}^{H} \mid o_t, \ell}\Big[\prod_{h} q(z_h \mid o_{t+h}, o_{t+h+1})\Big].
\end{equation}
The optimal flow under $\mathcal{L}_{\text{FM}}$ transports $\mathcal{N}(0,I)$ to $p^{*}$~\cite{lipman2022flow}, a Gaussian mixture over plausible futures that can capture action multimodality. At any timestep $\tau$, an estimate of the latent action is recovered as $ \hat{z}_{\tau,h} = x_{\tau,h} + (1-\tau)\,v_\theta(x_{\tau,h}, o_t, \ell)$. 

\mypara{Flow-Adaptive Posterior Distillation.}
The FM loss matches samples from the teacher posterior and preserves the stochasticity of possible latent actions. We further introduce a distillation term that provides a covariance-aware anchor to the high-probability region of the teacher posterior by aligning $\hat{z}_{\tau,h}$ with the posterior mean $\mu_h$ under the inverse-covariance metric $\Sigma_h^{-1}$. At early timesteps ($\tau \to 0$), where noisy velocity dominates $\hat{z}_{\tau,h}$, a constant penalty would bias the velocity field toward the posterior mean and hinder exploration of the transport path. We therefore weight the penalty by $\tau$, yielding the flow-adaptive distillation loss:
\begin{equation}
    \mathcal{L}_{\text{distill}} = \mathbb{E}_{\tau,\, x_{0,h},\, z_h} \!\left[ \frac{\tau}{2} \bigl\| \hat{z}_{\tau,h} - \mu_h \bigr\|_{\Sigma_h^{-1}}^2 \right].
\end{equation}

\subsection{Correspondence as Motion Supervision}
\label{sec:cms}

Latent action models typically rely on pixel reconstruction, which entangles the latent space with task-irrelevant content. Prior work factorizes the latent~\cite{bu2025univla,li2025latbot} or introduces optical flow~\cite{bu2025laof,bi2025motus}. Factorization requires a dedicated design to enforce the decomposition, while optical flow relies on an external estimator to generate motion targets. We instead leverage the cross-patch correspondence of the frozen DINOv2 encoder $\mathcal{E}$, whose stable cross-frame features expose motion as a feature-space residual.

\mypara{Correspondence-Derived Motion Field.}
For each chunk step $h$, we extract the patch tokens $f_{t+h}, f_{t+h+1} \in \mathbb{R}^{N \times d_\text{obs}}$ and last-block key vectors $k_{t+h}, k_{t+h+1} \in \mathbb{R}^{N \times d_k}$ from the frozen encoder $\mathcal{E}$. These keys act as cross-frame descriptors~\cite{oquab2023dinov2}, from which we form a softmax correspondence matrix $C_h \in \mathbb{R}^{N \times N}$ whose rows softly match current frame to the next:
\begin{equation}
    C_h = \mathrm{Softmax}\left(\frac{k_{t+h}\, k_{t+h+1}^\top}{\sqrt{d_k}}\right).
\end{equation}
Warping $f_{t+h+1}$ with $C_h$ yields a dense correspondence residual $\Delta f_h$:
\begin{equation}
    \tilde{f}_h = C_h\, f_{t+h+1}, \qquad \Delta f_h = \tilde{f}_h - f_{t+h}.
\end{equation}
Since DINOv2 features stay stable across adjacent frames~\cite{oquab2023dinov2}, $\tilde{f}_h$ and $f_{t+h}$ share identity content and differ mainly in motion. Their residual $\Delta f_h$ is a feature-space proxy for that motion, measuring how matched content changes once re-indexed onto the current patches.

\mypara{Latent-Conditioned Motion Predictor.}
We connect the latent action policy to the target $\Delta f_h$ with a motion predictor $\phi_{\text{mot}}$ that estimates a per-step residual over the chunk-anchor patch tokens $f_t$, conditioned on $\hat{Z}_\tau = \{\hat{z}_{\tau,h}\}_{h=0}^{H-1}$. We pass the latents via a causal self-attention, so each $\bar{z}_{\tau,h} = \mathrm{CausalAttn}_h(\hat{Z}_\tau)$ aggregates only from its predecessors, and then modulates $f_t$ via AdaLN:
\begin{equation}
    {[\gamma_h, \beta_h]} = \psi_{\text{mod}}(\bar{z}_{\tau,h}), \quad u_{h,n} = \mathrm{LN}(f_{t,n}) \odot (1 + \gamma_h) + \beta_h,
\end{equation}
where $\psi_{\text{mod}}$ is a shallow MLP. Since $f_t$ comes from the frozen $\mathcal{E}$, no gradient flows through it. By construction, $f_t$ carries the static scene structure shared across all $H$ steps, while $\bar{z}_{\tau,h}$ provides the per-step action signal. A patch-wise decoder $\mathcal{P}$ maps the modulated tokens $u_{h,n}$ to per-patch residuals $\Delta\hat{f}_{h,n} = \mathcal{P}(u_{h,n})$, supervised by:
\begin{equation}
    \mathcal{L}_{\text{motion}} = \frac{1}{N} \sum_{n=1}^{N} \left( 1 - \frac{\Delta\hat{f}_{h,n}^\top \Delta f_{h,n}}{\|\Delta\hat{f}_{h,n}\|\,\|\Delta f_{h,n}\|} \right).
\end{equation}
This supervision trains $\phi_{\text{mot}}$ as an action-conditioned latent dynamics model that predicts correspondence residuals from the current visual features and latent actions, encouraging the latent actions to encode motion dynamics. 
We jointly optimize the losses on action-free videos:
\begin{equation}
    \mathcal{L}_{\text{total}} = \mathcal{L}_{\text{FM}} + \lambda_{\text{d}} \mathcal{L}_{\text{distill}} + \lambda_{\text{m}} \mathcal{L}_{\text{motion}}.
\end{equation}
At inference, we obtain the final latent chunk $\hat{Z} = \{\hat{z}_h\}_{h=0}^{H-1}$ by integrating $v_\theta$ from $\tau{=}0$ to $1$ given $o_t$ and $\ell$, and discard the continuous LAM teacher and motion predictor. Algorithm~\ref{alg:stage1} summarizes the training process.

\begin{algorithm}[t]
\caption{Stage~I training of \ours{}.}
\label{alg:stage1}
\begin{algorithmic}[1]
\Require dataset $\mathcal{D}$, horizon $H$, weights $\lambda_{\text{d}}, \lambda_{\text{m}}$, learning rate $\eta$, frozen $\mathcal{E}$ and LAM $q$
\Repeat
    \State $\big(\{o_{t+j}\}_{j=0}^{H},\, \ell\big) \sim \mathcal{D}$,\enspace $\{f_{t+j},\, k_{t+j}\}_{j=0}^{H} \gets \mathcal{E}\big(\{o_{t+j}\}_{j=0}^{H}\big)$
    \State $(\mu_h, \Sigma_h) \gets q(o_{t+h}, o_{t+h+1})$,\enspace $z_h \sim \mathcal{N}(\mu_h, \Sigma_h)$ \Comment{teacher target}
    \State $C_h \gets \mathrm{Softmax}\big(k_{t+h}\, k_{t+h+1}^\top / \sqrt{d_k}\big)$,\enspace $\Delta f_h \gets C_h\, f_{t+h+1} - f_{t+h}$ \Comment{motion target}
    \State $\tau \sim \mathcal{U}(0,1)$,\enspace $x_{0,h} \sim \mathcal{N}(0, I)$,\enspace $x_{\tau,h} \gets (1-\tau)\,x_{0,h} + \tau\, z_h$
    \State $\hat{v}_h \gets v_\theta(x_{\tau,h},\, o_t,\, \ell)$,\enspace $\hat{z}_{\tau,h} \gets x_{\tau,h} + (1-\tau)\,\hat{v}_h$ \Comment{student forward}
    \State $\{\Delta\hat{f}_h\}_{h=0}^{H-1} \gets \phi_{\text{mot}}(\hat{Z}_\tau,\, f_t)$ \Comment{motion prediction}
    \State $\theta \gets \theta - \eta\, \nabla_\theta \big(\mathcal{L}_{\text{FM}} + \lambda_{\text{d}} \mathcal{L}_{\text{distill}} + \lambda_{\text{m}} \mathcal{L}_{\text{motion}}\big)$
\Until{converged}
\end{algorithmic}
\end{algorithm}

\subsection{Embodiment-Specific Action Decoding}
\label{sec:lad}

To deploy the pretrained latent policy on a target embodiment, a decoder is required to map the latent chunk to its action space. We train this decoder from scratch and finetune the latent policy on the target's labeled actions.
Let $p_t \in \mathbb{R}^{d_p}$ denote the proprioceptive state and $f_t \in \mathbb{R}^{N \times d_\text{obs}}$ the chunk-anchor patch tokens reused from \secref{sec:cms}. The decoder $\phi_{\text{dec}}$ produces the embodiment-specific action chunk $\hat{A} = \{\hat{a}_h\}_{h=0}^{H-1}$ with $\hat{a}_h \in \mathbb{R}^{d_a}$ from $\hat{Z} = \{\hat{z}_h\}_{h=0}^{H-1}$, conditioned on $p_t$ and $f_t$ along distinct pathways. As a global hardware descriptor, $p_t$ passes through an MLP $\psi_{\text{prop}}$ and modulates every step feature-wise via AdaLN, conditioning the chunk uniformly on the target hardware. As a spatially structured sequence, $f_t$ is instead fed into cross-attention, allowing each step to attend to relevant regions. We augment $\hat{Z}$ with a learned positional embedding, compose the streams through $L$ stacked layers, and read out $\hat{A}$ with an MLP head. We optimize the decoder by a chunk-level mean-squared error:
\begin{equation}
    \mathcal{L}_{\text{act}} = \frac{1}{H\, d_a} \sum_{h=0}^{H-1} \left\| \hat{a}_h - a_h \right\|_2^2,
\end{equation}
where $d_a$ is the action dimension. Action labels enter only through $\mathcal{L}_{\text{act}}$, so Stage~I stays action-free and the latent policy serves as a shared initialization that Stage~II adapts to each embodiment.

\section{Experiment}
\label{sec:exp}

\mypara{Implementation Details.}
Our latent policy is a $16$-block flow-matching Transformer predicting an $H{=}16$ latent chunk of dimension $d_z{=}32$, conditioned on frozen DINOv2~\cite{oquab2023dinov2} and SigLIP2~\cite{tschannen2025siglip} visual features and Flan-T5~\cite{chung2024scaling} text embeddings. The motion supervision features $f_t$ and keys $k_t$ come from DINOv2, and the LAM teacher is DreamDojo~\cite{gao2026dreamdojo}. Excluding the frozen LAM, the model totals $1.1$B parameters, of which only $590$M are trainable, since the backbones are frozen and only the latent policy, motion predictor, and the $8$-block action decoder are learned. Stage~I pretrains the latent policy and motion predictor under $\mathcal{L}_{\text{total}}$ on label-free training data of human (EgoDex~\cite{hoque2025egodex}) and multi-robot (RoboCasa-GR1~\cite{bjorck2025gr00t}, Fractal~\cite{collaboration2023open}, BridgeData~V2~\cite{walke2023bridgedata}, RoboCasa365~\cite{nasiriany2026robocasa365}, DreamDojo-GR1~\cite{gao2026dreamdojo}) videos for $250$k steps at batch size $128$. Stage~II trains an action decoder from scratch and finetunes the policy under $\mathcal{L}_{\text{act}}$ for each target embodiment. Refer to~\secref{sec:sup-impl} for more architectural and training details.

\begin{figure*}[t]
    \centering
    \includegraphics[width=0.96\linewidth]{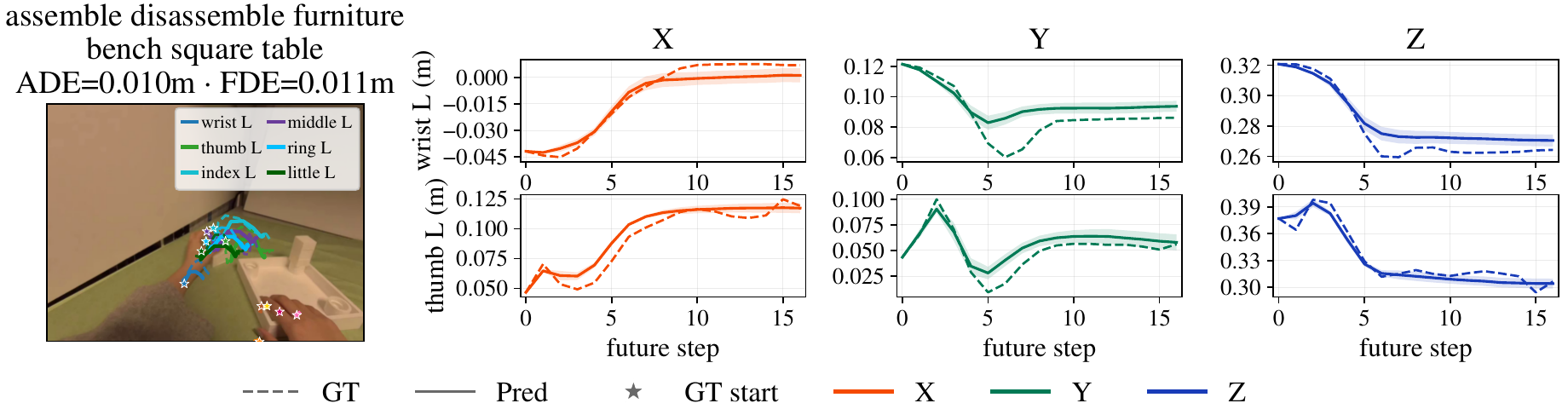}
    \caption{Qualitative hand pose forecasting on EgoDex. \emph{Left} overlays the predicted $1.6$\,s keypoint trajectories on the egocentric frame. \emph{Right} plots each keypoint's $X$/$Y$/$Z$ position over the horizon.}
    \label{fig:egodex_traj}
\end{figure*}

\mypara{Evaluation Benchmarks.}
We evaluate \ours{} on three benchmarks, \emph{EgoDex}~\cite{hoque2025egodex} for bimanual hand pose forecasting, \emph{SimplerEnv}~\cite{li2024evaluating} (\emph{WidowX} and \emph{Google Robot}) for real-to-sim single-arm manipulation, and \emph{RoboCasa-GR1}~\cite{bjorck2025gr00t} for humanoid tabletop manipulation. The three benchmarks span different action spaces, a $48$-D bimanual hand pose for EgoDex, a $7$-DoF single arm for SimplerEnv, and a $29$-D humanoid joint space for RoboCasa-GR1. We report Average and Final Displacement Error (ADE\,/\,FDE) for EgoDex and success rate for the SimplerEnv and RoboCasa-GR1 benchmarks. See~\secref{sec:sup-benchmarks} for data statistics and evaluation protocols. We also deploy \ours{} on an xArm7, an embodiment unseen in Stage~I pretraining, for three tabletop tasks, reporting success rate over $50$ rollouts per task.

\begin{table*}[t]
    \centering
    \begin{minipage}[t]{0.49\linewidth}
        \centering
        \resizebox{0.97\linewidth}{!}{
    \begin{tabular}{y{25mm}|x{8mm}x{8mm}x{8mm}|x{8mm}x{8mm}x{8mm}}
        \toprule
        \multirow{2}{*}{\textbf{Methods}}
        & \multicolumn{3}{c|}{\textbf{ADE} ($m$)$\downarrow$}
        & \multicolumn{3}{c}{\textbf{FDE} ($m$) $\downarrow$} \\
        \cmidrule(lr){2-4} \cmidrule(lr){5-7}
        & $S{=}1$ & $S{=}5$ & $S{=}10$ & $S{=}1$ & $S{=}5$ & $S{=}10$ \\
        \midrule
        Dec + BC & 0.039 & 0.039 & 0.039 & 0.053 & 0.053 & 0.053 \\
        Dec + DDPM & 0.054 & 0.041 & 0.039 & 0.071 & 0.050 & 0.044 \\
        Dec + FM & 0.052 & 0.041 & 0.038 & 0.071 & 0.051 & 0.044 \\
        \midrule
        EncDec + BC & 0.040 & 0.040 & 0.040 & 0.053 & 0.053 & 0.053 \\
        EncDec + DDPM & 0.058 & 0.044 & 0.041 & 0.074 & 0.052 & 0.046 \\
        EncDec + FM & 0.059 & 0.045 & 0.041 & 0.077 & 0.054 & 0.047 \\
        \midrule
        \rowcolor{ourscolor} \textbf{\ours} & \textbf{0.034} & \textbf{0.031} & \textbf{0.029} & \textbf{0.047} & \textbf{0.040} & \textbf{0.038}\\
        \bottomrule
    \end{tabular}
}

        \captionof{table}{EgoDex hand pose forecasting results. ADE and FDE are best-of-$S$ errors in meters.}
        \label{tab:egodex}
    \end{minipage}\hfill
    \begin{minipage}[t]{0.49\linewidth}
        \centering

\resizebox{0.96\linewidth}{!}{
    \begin{tabular}{y{27mm}|x{10mm}x{12mm}x{10mm}|x{12mm}}
        \toprule
        \textbf{Tasks} & GR00T-N1.6 & StarVLA-$\pi$ & Qwen3-DiT & \textbf{\ours{}} \\
        \midrule
        BottleToCabinet & 51.5 & 26.0 & 50.0 & 42.0 \\
        CanToDrawer& 13.0 & 62.0 & 48.0 & 78.0 \\
        CupToDrawer & 8.5 & 42.0 & 42.0 & 24.0 \\
        MilkToMicrowave & 14.0 & 50.0 & 38.0 & 44.0 \\
        PotatoToMicrowave& 41.5 & 42.0 & 18.0 & 34.0 \\
        WineToCabinet & 16.5 & 32.0 & 28.0 & 34.0 \\
        \midrule
        \textbf{Avg} & 24.2 & 42.3 & 37.3 & \textbf{42.7} \\

        \bottomrule
    \end{tabular}
}

        \captionof{table}{RoboCasa-GR1 success rate (\%). Avg is over the six tasks.}
        \label{tab:gr1_detail}
    \end{minipage}
\end{table*}

\begin{table*}[t]
    \centering
    \resizebox{0.98\linewidth}{!}{
    \begin{tabular}{y{23mm}|x{10mm}|x{14mm}x{14mm}x{24mm}x{24mm}|x{10mm}}
        \toprule
        \textbf{Models} & \textbf{Params}
            & \shortstack{Put Spoon \\ on Towel} & \shortstack{Put Carrot \\ on Plate} & \shortstack{Stack Green Block \\ on Yellow Block} & \shortstack{Put Eggplant \\ in Yellow Basket} & \textbf{Avg} \\
        \midrule
        RT-1-X     & 35M  & 0.0  & 4.2  & 0.0  & 0.0  & 1.1  \\
        DINOv2 + DP    & 86M    & 0.0  & 8.0  & 0.0  & 26.0 & 8.5  \\
        Octo-Base           & 93M  & 15.8 & 12.5 & 0.0  & 41.7 & 17.5 \\
        OpenVLA           & 7B   & 8.3  & 4.2  & 0.0  & 45.8 & 14.6 \\
        $\pi_0$            & 3B   & 29.1 & 0.0  & 16.6 & 62.5 & 27.1 \\
        GR00T-N1         & 2B   & 62.5 & 45.8 & 16.7 & 20.8 & 36.5 \\
        RoboVLMs          & 2B   & 45.8 & 20.8 & 4.2  & 79.2 & 37.5 \\
        villa-X            & 3B   & 48.3 & 24.2 & 19.2 & 71.7 & 40.8 \\
        UniVLA              & 7B   & 52.8 & 55.6 & 2.8  & 80.6 & 47.9 \\
        mimic-video         & 2B   & 37.5 & 37.5 & 12.5 & 100.0 & 46.9 \\
        \cmidrule(lr){1-7}
        \rowcolor{ourscolor} \textbf{\ours{(Ours)}}   & 1.1B   & 58.3 & 54.2 & 12.5 & 83.3 & \textbf{52.1} \\
        \bottomrule
    \end{tabular}
}

    \vspace{1pt}
    \resizebox{0.98\linewidth}{!}{
    \begin{tabular}{y{23mm}|x{10mm}|x{8mm}x{8mm}x{17mm}|x{10mm}|x{8mm}x{8mm}x{17mm}|x{10mm}}
        \toprule
        \multirow{2}{*}{\textbf{Models}} & \multirow{2}{*}{\textbf{Params}}
            & \multicolumn{4}{c|}{\textit{Visual Matching (VM)}}
            & \multicolumn{4}{c}{\textit{Variant Aggregation (VA)}} \\
        \cmidrule(lr){3-6} \cmidrule(lr){7-10}
            &
            & \shortstack{Pick \\ Can} & \shortstack{Move \\ Near} & \shortstack{Open/Close \\ Drawer} & \textbf{Avg}
            & \shortstack{Pick \\ Can} & \shortstack{Move \\ Near} & \shortstack{Open/Close \\ Drawer} & \textbf{Avg} \\
        \midrule
        RT-1-X     & 35M  & 56.7 & 31.7 & 59.7 & 49.4 & 49.0 & 32.3 & 29.4 & 36.9 \\
        DINOv2 + DP    & 86M    & 31.0 & 5.0  & 60.0 & 32.0 & -    & -    & -    & -    \\
        Octo-Base           & 93M  & 17.0 & 4.2  & 22.7 & 14.6 & 0.6    & 3.1    & 1.1   & 1.6  \\
        OpenVLA           & 7B   & 18.0 & 56.3 & 63.0 & 45.8 & 60.8 & 67.7 & 28.8 & 52.4 \\
        $\pi_0$            & 3B   & 72.7 & 65.3 & 38.3 & 58.8 & 75.2 & 63.7 & 25.6 & 54.8 \\
        GR00T-N1         & 2B   & 47.0 & 70.0 & 18.1 & 45.0 & 78.8 & 62.5 & 13.2 & 51.5 \\
        GR00T-N1.5       & 2B   & 51.7 & 54.0 & 27.8 & 44.5 & 69.3 & 68.7 & 35.8 & \textbf{57.9} \\
        villa-X            & 3B   & 81.7 & 55.4 & 38.4 & 58.5 & -    & -    & -    & -    \\
        DeFI       & 2.5B    & 54.2 & 60.7 & 38.6 & 51.2 & -    & -    & -    & -    \\
        \cmidrule(lr){1-10}
        \rowcolor{ourscolor} \textbf{\ours{(Ours)}}   & 1.1B   & 68.3 & 67.1 & 43.5 & \textbf{59.6} & 84.3 & 51.5 & 35.5 & 57.1 \\
        \bottomrule
    \end{tabular}
}

    \caption{Per-task success rate (\%) on SimplerEnv~\cite{li2024evaluating}. \emph{Top}: WidowX. \emph{Bottom}: Google Robot. Baselines are RT-1-X~\cite{collaboration2023open}, DINOv2\,+\,DP~\cite{dong2026capturing}, Octo-Base~\cite{team2024octo}, OpenVLA~\cite{kim2024openvla}, $\pi_0$~\cite{black2024pi_0}, GR00T-N1 and N1.5~\cite{bjorck2025gr00t}, RoboVLMs~\cite{liu2025towards}, villa-X~\cite{chen2025villa}, UniVLA~\cite{bu2025univla}, mimic-video~\cite{pai2025mimic}, and DeFI~\cite{zhang2026disentangled}.}
    \label{tab:simplerenv}
\end{table*}

\subsection{Main Results}

\noindent\textbf{Hand Pose Forecasting.}
\tabref{tab:egodex} reports our results on EgoDex, comparing against BC, DDPM, and FM variants of the decoder-only (Dec) and encoder-decoder (EncDec) architectures from X-IL~\cite{jia2025x}.
Best-of-$S$ reports the lowest error across $S$ samples, so the deterministic BC baselines stay flat in $S$ while the stochastic diffusion and flow-matching baselines improve with it. Our method posts $0.029$\,/\,$0.038$\,m ADE\,/\,FDE at $S{=}10$, and its relative ADE margin over the strongest baseline widens from $13\%$ at $S{=}1$ to $24\%$ at $S{=}10$. Since this gap also holds against the flow-matching baselines, it stems from modeling actions in a latent space, not from flow matching itself. \figref{fig:egodex_traj} plots a predicted trajectory whose keypoints track the ground truth and diverge only late in the horizon, indicating that the policy captures fine finger-level motion.

\noindent\textbf{Real-to-Sim Evaluation on SimplerEnv.}
In~\tabref{tab:simplerenv}, we group the baselines into policies optimized end-to-end on labeled robot data (e.g., $\pi_0$~\cite{black2024pi_0}, GR00T-N1~\cite{bjorck2025gr00t}) and methods that prepend an action-free pretraining stage (UniVLA~\cite{bu2025univla}, mimic-video~\cite{pai2025mimic}). Since AdaWorld and DreamDojo are world models rather than standalone control policies, we use such an encoder as our frozen LAM teacher rather than as a policy baseline. CLAM learns a separate policy for each task, whereas \ours{} learns one language-conditioned policy for all tasks. On \emph{WidowX}, \ours{} scores $52.1\%$ while optimizing only $590$M of its $1.1$B parameters, exceeding UniVLA and mimic-video by $4.2$ and $5.2$, respectively. We attribute this advantage to modeling latent actions as a continuous flow rather than a discrete codebook, which retains the multimodal structure of plausible futures. \emph{Google Robot} instead probes robustness to visual variation. \ours{} achieves $59.6\%$ under Visual Matching and $57.1\%$ under the perturbed Variant Aggregation protocol, a degradation of only $2.5$. We trace this stability to our motion predictor, since a latent grounded in DINOv2 correspondences encodes motion structure largely invariant to appearance.

\noindent\textbf{Manipulation on RoboCasa-GR1.}
\tabref{tab:gr1_detail} compares \ours{} on $6$ articulated multi-step RoboCasa-GR1 tasks against GR00T-N1.6~\cite{bjorck2025gr00t}, StarVLA-$\pi$~\cite{community2026starvla}, and Qwen3DiT~\cite{ma2026dit4dit}. Unlike these, \ours{} adapts its action-free pretrained latent to the humanoid rather than learning the policy end-to-end on labeled actions. Aggregated over the six tasks, it manages a comparable $42.7\%$ average, a low-cost transfer to a high-DoF embodiment.

\begin{figure*}[t]
    \centering
    \includegraphics[width=0.98\linewidth]{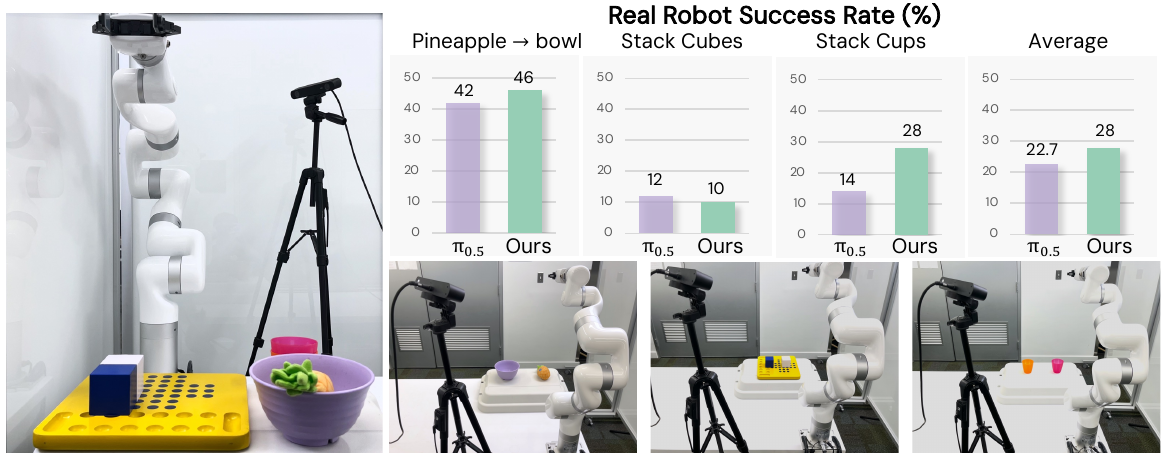}
    \caption{Real-world setup and results. \emph{Left}, the single-camera xArm7 tabletop workspace. \emph{Right}, per-task and average success rate over $50$ rollouts. See the supplementary video for details.}
    \label{fig:realrobot}
\end{figure*}

\noindent\textbf{Real-World Deployment.} We deploy \ours{} on a physical xArm7 with a ZED~2i (\emph{left view}) camera, comparing with $\pi_{0.5}$~\cite{intelligence2025pi_} on three tabletop tasks (\figref{fig:realrobot}). Following our train recipe, we collect $50$ teleoperated demonstrations per task, fit an $8$-DoF joint-space decoder from scratch, and finetune the latent policy under $\mathcal{L}_{\text{act}}$. The three tasks share a pick-and-place structure but escalate in placement precision, from a tolerant drop (\texttt{Pineapple$\rightarrow$Bowl}) through aligned nesting (\texttt{Stack Cups}) to a tight stack against soft-contact rebound (\texttt{Stack Cubes}). Across $50$ rollouts evaluation per task, \ours{} averages $28.0\%$ against $\pi_{0.5}$'s $22.7\%$. Results track this precision ordering and localize where the latent helps. Because of the task-intrinsic difficulty, both methods struggle with cube stacking. In contrast, \ours{} doubles $\pi_{0.5}$'s success rate on cup stacking, benefiting from a continuous latent that preserves the fine motor distinctions. On an unseen compositional instruction (``pick up the red cup and put it into the bowl''), \ours{} succeeds on $23$ of $50$ rollouts, recombining learned grounding and skills rather than replaying memorized trajectories. For more real-world analysis, inference latency, and per-task diagnostics see~\secref{sec:sup-realrobot}.

\subsection{Analysis and Ablation}
\label{sec:analysis}

\begin{figure}[t]
    \centering
    \begin{minipage}[b]{0.51\linewidth}
        \centering
        \includegraphics[width=\linewidth]{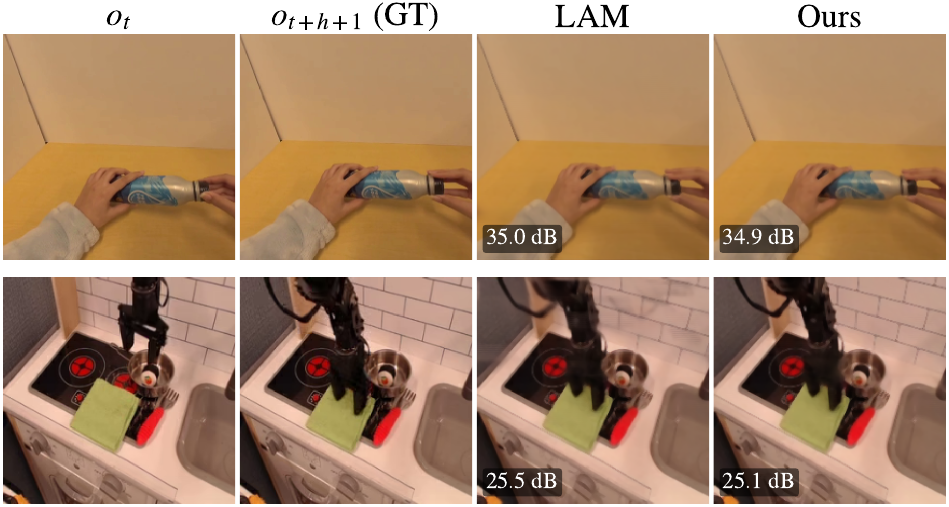}
    \end{minipage}\hfill
    \begin{minipage}[b]{0.47\linewidth}
        \centering
        \resizebox{0.99\linewidth}{!}{
    \begin{tabular}{y{40mm}|x{18mm}|x{20mm}}
        \toprule
        \multirow{2}{*}{\textbf{Variant}}
            & \textbf{EgoDex} ($m$)
            & \textbf{WidowX} ($\%$) \\
        \cmidrule(lr){2-2} \cmidrule{3-3}
            & FDE $\downarrow$ & SR (3 tasks) $\uparrow$ \\
        \midrule
        \rowcolor{ourscolor} \textbf{\ours} (full)              & $\textbf{0.0384}$ & $\textbf{44.4}$ \\
        \midrule
        \multicolumn{3}{l}{\textit{Stage~I objective}} \\
        \quad $\mathcal{L}_{\text{FM}}$ only                    & $0.0391$ & $40.3$ \\
        \quad Constant weight in $\mathcal{L}_{\text{distill}}$  & $0.0388$ & $38.9$ \\
        \midrule
        \multicolumn{3}{l}{\textit{Stage~I training data}} \\
        \quad EgoDex only & $0.0392$ & $34.7$ \\
        \quad Robot only  & $0.0469$ & $41.7$ \\
        \midrule
        \multicolumn{3}{l}{\textit{Stage~I teacher (robot-only data, compare with Robot only)}} \\
        \quad UniVLA VQ teacher & $0.0482$ & $29.2$ \\
        \bottomrule
    \end{tabular}
}

    \end{minipage}\\
    \begin{minipage}[t]{0.51\linewidth}
        \centering
        \captionof{figure}{Latent action quality probing. Each row shows $o_t$ for reference, the GT $o_{t+h+1}$, and reconstructions from the teacher latent and from our $\hat{z}_h$.}
        \label{fig:z_pred_quality}
    \end{minipage}\hfill
    \begin{minipage}[t]{0.47\linewidth}
        \centering
        \captionof{table}{Component ablations. Every variant changes Stage~I only and shares the same Stage~II recipe and protocol.}
        \label{tab:ablation}
    \end{minipage}
\end{figure}

\begin{figure}[t]
    \centering
    \includegraphics[width=0.99\linewidth]{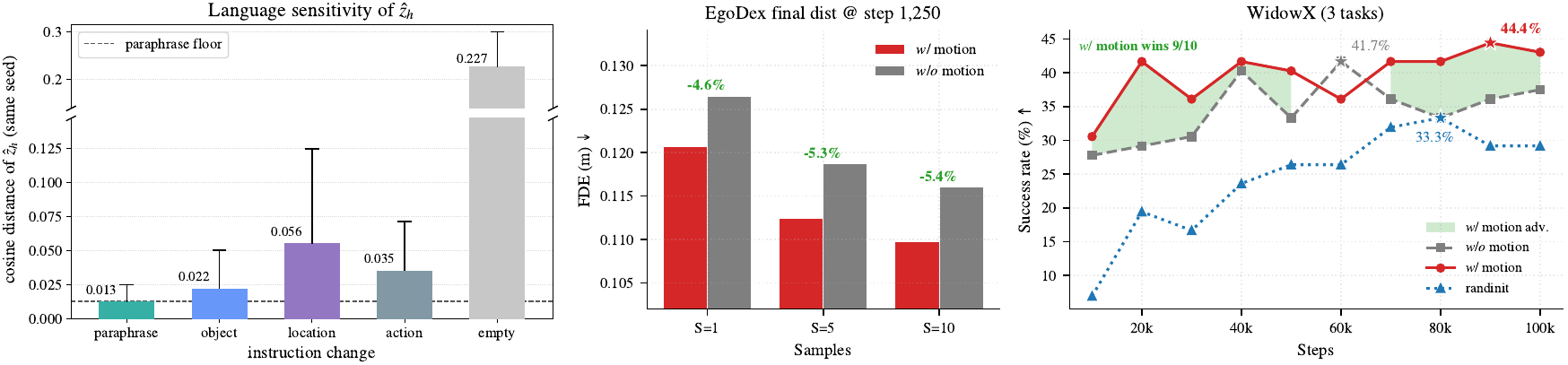}
    \caption{Language sensitivity (\emph{left}), motion supervision, and pretraining (\emph{middle \& right}) ablations.}
    \label{fig:motion}
\end{figure}

\noindent\textbf{Latent Action Probing.}
We probe learned latent $\hat{z}_h$ along three axes, each isolating a property the policy needs. \emph{Teacher fidelity.} The LAM teacher forms posterior from $(o_{t+h}, o_{t+h+1})$ while our policy predicts $\hat{z}_h$ from $o_t$ alone. Decoded from $o_{t+h}$ by the \emph{same} frozen LAM decoder~\cite{gao2026dreamdojo}, $\hat{z}_h$ reconstructs $o_{t+h+1}$ close to the teacher's (\figref{fig:z_pred_quality}, where $o_t$ is shown for reference). The policy thus recovers the teacher's predictive content without observing the future. \begin{wrapfigure}{r}{0.40\linewidth}
    \vspace{-1.0\baselineskip}
    \centering
    \includegraphics[width=\linewidth]{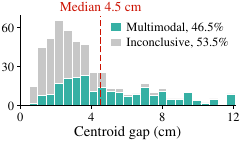}
    \vspace{-1.5\baselineskip}
    \caption{Latent multimodality.}
    \label{fig:multimodal}
    \vspace{-1.2\baselineskip}
\end{wrapfigure}
\emph{Language sensitivity.} The LAM teacher takes no instruction, so understanding language is a capability the policy must learn. The same frame recurs under different instructions with different teacher targets, and $\ell$ is the only input that separates them. Holding $o_t$ fixed while varying the instruction, $\hat{z}_h$ stays near the floor under paraphrase or object swaps but shifts $2.7\times$ and $4.3\times$ under verb or location changes and $17.5\times$ when language is removed (\figref{fig:motion}), tracking what motion and where. \emph{Multimodality.} To evaluate latent diversity, we decode $64$ latents per observation using the frozen action decoder. For $46.5\%$ of $512$ EgoDex observations, the predicted endpoints split into two distinct clusters (median separation $4.5$\,cm), compared to only $2$ of $100$ unimodal controls. This confirms that our model captures genuinely multimodal action distributions rather than unimodal averages (\figref{fig:multimodal}, \secref{sec:sup-multimodal}).

\noindent\textbf{Ablation.}
We vary Stage~I while keeping the Stage~II training recipe and evaluation protocol unchanged (\tabref{tab:ablation} \& \figref{fig:motion}). \emph{Stage~I pretraining.} A randomly initialized latent policy, adapted by the same Stage~II, peaks at $33.3\%$ WidowX SR against the full model's $44.4\%$ (\figref{fig:motion}), confirming that Stage~I pretraining drives the downstream gain. \emph{Distillation objective.}
The $\mathcal{L}_{\text{FM}}$-only variant reaches $40.3\%$ WidowX SR, our $\tau$-weighting raises it to $44.4\%$, and a constant weight drops it to $38.9\%$. This constant weight enforces the alignment to the posterior mean even when the latent is still noise, which is why it ends up worse than no distillation at all. Both ablated latents also reconstruct the future at lower PSNR (\secref{sec:sup-recon}). \emph{Training data.} Pretraining on EgoDex or robot videos alone trails the mixture by $9.7$ and $2.7$ WidowX and is also worse on EgoDex itself, so human and robot data contribute complementary priors. \emph{Teacher representation.} Replacing the continuous teacher~\cite{gao2026dreamdojo} with the VQ LAM released by UniVLA~\cite{bu2025univla}, which encodes each frame transition into four tokens from a $16{\times}128$ codebook, lowers WidowX SR from $41.7$ to $29.2$ and raises EgoDex FDE from $0.0469$ to $0.0482$ under the robot-only data setting that both runs share. \emph{Motion supervision.} Adding $\mathcal{L}_{\text{motion}}$ reduces early-training EgoDex FDE by $4.6$ to $5.4$\%, and the motion-supervised variant wins $9$ of $10$ WidowX checkpoints (\figref{fig:motion}), so the correspondence target grounds the latent in dynamics throughout training rather than merely warming it up.

\section{Conclusion and Limitations}
\label{sec:conclusion}

We presented \ours{}, a unified latent action policy for cross-embodiment manipulation, learned from action-free videos. It models the latent action as a continuous flow-matching policy distilled from a frozen LAM teacher under a flow-adaptive posterior objective, anchors the latent in motion via correspondence-based supervision in a frozen ViT, and grounds it on each robot with a separate action decoder and a light policy finetune. Across hand pose forecasting, single-arm and humanoid manipulation, \ours{} compares favorably with prior latent-action and VLA baselines while training far fewer parameters than the larger ones. We hope this continuous, motion-grounded latent offers a reusable, embodiment-agnostic interface for learning manipulation from large-scale action-free videos across human and robots.

\ours{} has several limitations that scope our claims and point to future work. At $1.1$B parameters with $590$M trainable, it keeps visual and language backbones frozen for efficiency, rather than co-training them end-to-end~\cite{community2026starvla}, so scaling or unfreezing them to gain perceptual capacity is a natural step. Reliance on a single camera may limit performance under occlusion, so how to integrate complementary observations, such as a wrist view, remains open. Scaling the human corpus beyond EgoDex offers a further route to a stronger shared latent, since our ablation shows human and robot videos contribute complementary priors.

\subsection*{Acknowledgments}
We thank Siyu Li (University of Pennsylvania) for helpful discussions.

\renewcommand{\bibfont}{\small}
\setlength{\bibsep}{4pt plus 1pt minus 1pt}
\bibliography{reference}
\bibliographystyle{iclr2027_conference}

\appendix
\clearpage

\section*{Overview}
\addcontentsline{toc}{section}{Overview}
This supplementary material extends the main paper in four parts. \secref{sec:sup-impl} specifies the model architecture as well as the training and inference details. \secref{sec:sup-benchmarks} describes the EgoDex, SimplerEnv, and RoboCasa-GR1 evaluation benchmarks and their protocols. \secref{sec:sup-realrobot} details the real-robot setup, the task suite, qualitative rollouts, and a per-task analysis. \secref{sec:sup-analysis} reports additional analysis, covering latent probing, language sensitivity, hand-pose forecasts, and latent reconstructions. \tabref{tab:sup_notation} collects the notation used in this paper.

\begin{table}[!ht]
    \centering
    \small
    \begin{tabular}{y{45mm} y{73mm}}
        \toprule
        \textbf{Symbol} & \textbf{Meaning} \\
        \midrule
        $t$, $H$, $h$ & time step, chunk horizon and chunk index \\
        $N$, $n$ & number of patch tokens per frame and patch index \\
        $o_t$, $\ell$ & current observation frame and language instruction \\
        $\tau$ & flow timestep \\
        $z_h\in\mathbb{R}^{d_z}$ & latent action of the transition $(o_{t+h},o_{t+h+1})$, sampled from the LAM teacher \\
        $\mu_h$, $\Sigma_h$ & teacher posterior mean and diagonal covariance \\
        $\pi(\cdot\mid o_t,\ell)$, $v_\theta$ & latent policy and the velocity field \\
        $x_{0,h}$, $x_{\tau,h}$ & initial noise and its linear interpolant toward $z_h$ at flow timestep $\tau$ \\
        $p^{*}$ & teacher posterior marginalized over the unobserved future, the target of $\mathcal{L}_{\text{FM}}$ \\
        $\hat{z}_{\tau,h}$ & estimate of the latent action at flow timestep $\tau$, with $\hat{z}_h$ its endpoint at $\tau{=}1$ \\
        $\mathcal{E}$ & frozen DINOv2 encoder \\
        $f_{t+h}\in\mathbb{R}^{N\times d_{\text{obs}}}$, $k_{t+h}\in\mathbb{R}^{N\times d_k}$ & patch tokens and last-block keys of frame $o_{t+h}$ from $\mathcal{E}$ \\
        $f_t$ & patch tokens of the current frame ($f_{t+h}$ at $h{=}0$), the chunk anchor shared by all $H$ steps \\
        $C_h\in\mathbb{R}^{N\times N}$, $\tilde{f}_h$ & softmax correspondence matrix from the current frame to the next, and the next-frame tokens it warps onto the current patches \\
        $\Delta f_h$ & correspondence residual between the warped and current tokens, the motion target \\
        $\phi_{\text{mot}}$ & motion predictor \\
        $\bar{z}_{\tau,h}$ & latent aggregated causally over steps up to $h$ \\
        $\psi_{\text{mod}}$ & shallow MLP producing the AdaLN modulation \\
        $\gamma_h$, $\beta_h$ & AdaLN scale and shift \\
        $u_{h,n}$ & anchor token $n$ modulated for step $h$ \\
        $\mathcal{P}$ & patch-wise decoder \\
        $\Delta\hat{f}_{h,n}$ & predicted residual at patch $n$ \\
        $p_t\in\mathbb{R}^{d_p}$ & proprioceptive state \\
        $\phi_{\text{dec}}$, $\psi_{\text{prop}}$ & action decoder and the MLP embedding proprioception \\
        $\hat{A}$, $\hat{a}_h$, $a_h\in\mathbb{R}^{d_a}$ & predicted action chunk, its step-$h$ action and the ground-truth action \\
        $\mathcal{L}_{\text{FM}}$, $\mathcal{L}_{\text{distill}}$, $\mathcal{L}_{\text{motion}}$, $\mathcal{L}_{\text{total}}$ & Stage~I flow-matching, distillation and motion losses, and their weighted sum \\
        $\lambda_{\text{d}}$, $\lambda_{\text{m}}$ & weights of the distillation and motion losses in $\mathcal{L}_{\text{total}}$ \\
        $\mathcal{L}_{\text{act}}$ & Stage~II action loss \\
        \bottomrule
    \end{tabular}
    \caption{Notation used in \secref{sec:method}.}
    \label{tab:sup_notation}
\end{table}

\section{Model Architecture and Training Details}
\label{sec:sup-impl}

\subsection{Model Architecture}
Our latent action policy is a $16$-block flow-matching~\cite{lipman2022flow} Transformer that predicts an $H{=}16$ action chunk of latent dimension $d_z{=}32$ in a single forward pass. It conditions on the flow timestep with AdaLN~\cite{perez2018film,peebles2023scalable} and on the current observation frame $o_t$ and instruction tokens via per-block cross-attention. Two complementary visual encoders supply the visual conditioning, DINOv2-base (B)~\cite{oquab2023dinov2} for fine-grained spatial features and SigLIP2-base (B)~\cite{tschannen2025siglip} for vision-language semantics, and the frozen Flan-T5-large (L)~\cite{chung2024scaling} backbone embeds the instruction. Both visual encoders consume a $224{\times}224$ single-view RGB image. DINOv2-B (patch size $14$) yields $N{=}256$ patch tokens of width $d_{\text{obs}}{=}768$, and SigLIP2-B (patch size $16$) yields $196$ tokens, each linearly projected to the policy with width $1024$. The motion supervision reuses the DINOv2 patch tokens and its last-block key vectors of dimension $d_k{=}768$. The motion predictor reads the latent chunk via a $4$-head causal self-attention, modulates the frozen anchor features by AdaLN, and decodes patch-wise residuals with a two-layer MLP. This predictor is used only in Stage~I and discarded for Stage~II and inference. The embodiment-specific action decoder is an $8$-block Transformer which conditions on proprioception via AdaLN and on the anchor patch tokens with cross-attention. Only the latent policy, motion predictor, and action decoder are trainable ($590$M of the $1.1$B total parameters), while the visual and language backbones stay frozen. The $1.1$B excludes the frozen DreamDojo LAM, whose $404.8$M encoder provides the teacher posterior. \tabref{tab:sup_hparams} lists the architecture and optimization settings.

\begin{table*}[t]
    \centering
    {\small
\begin{tabular}{y{54mm} x{34mm}}
    \toprule
    \textbf{Component / Hyperparameter} & \textbf{Value} \\
    \midrule
    \multicolumn{2}{l}{\textit{Architecture}} \\
    \quad Latent policy: blocks\,/\,width\,/\,heads  & $16$\,/\,$1024$\,/\,$16$ \\
    \quad Action decoder: blocks\,/\,width\,/\,heads & $8$\,/\,$768$\,/\,$12$ \\
    \quad MLP ratio                           & $4$ \\
    \quad Action decoder dropout              & $0.1$ \\
    \quad Latent dimension $d_z$              & $32$ \\
    \quad Action chunk horizon $H$            & $16$ \\
    \quad Observation history                 & $1$ frame \\
    \quad Spatial visual encoder              & DINOv2-base \\
    \quad Semantic visual encoder             & SigLIP2-base \\
    \quad Text encoder                        & Flan-T5-large \\
    \quad Total\,/\,trainable parameters      & $1.1$B\,/\,$590$M \\
    \midrule
    \multicolumn{2}{l}{\textit{Stage~I pretraining (action-free)}} \\
    \quad Optimizer                           & AdamW \\
    \quad Learning-rate schedule              & cosine, $5\%$ warmup \\
    \quad Peak learning rate                  & $1{\times}10^{-4}$ \\
    \quad Weight decay                        & $0.01$ \\
    \quad Batch size                          & $128$ ($64{\times}2$ GPUs) \\
    \quad Training steps                      & $250$k \\
    \quad Precision                           & bf16-mixed \\
    \quad Gradient clip                       & $1.0$ \\
    \quad EMA decay                           & $0.9999$ \\
    \quad Distillation weight $\lambda_{\text{d}}$ & $0.05$ \\
    \quad Motion weight $\lambda_{\text{m}}$  & $0.5$ \\
    \midrule
    \multicolumn{2}{l}{\textit{Stage~II adaptation (per embodiment)}} \\
    \quad Action decoder initialization       & from scratch \\
    \quad Action decoder learning rate        & $1{\times}10^{-4}$ \\
    \quad Latent policy finetune learning rate & $1{\times}10^{-5}$ \\
    \quad Batch size (effective)              & $256$ ($16$ for the real robot) \\
    \quad Latent flow integration             & $4$-step Euler \\
    \quad Per-embodiment $d_a$, $d_p$, steps  & see \tabref{tab:sup_embodiments} \\
    \bottomrule
\end{tabular}%
}

    \caption{Architecture and optimization settings of \ours{}.}
    \label{tab:sup_hparams}
\end{table*}

\subsection{LAM Teacher and Latent Distillation}
The teacher is a frozen DreamDojo~\cite{gao2026dreamdojo} latent action encoder ($24$ blocks, width $1024$, patch size $16$) that maps each transition $(o_{t+h}, o_{t+h+1})$ to a diagonal-Gaussian posterior $\mathcal{N}(\mu_h, \Sigma_h)$ over the $d_z{=}32$ latent. Each transition is a consecutive frame pair at each corpus's native rate (stride $1$), namely $5$~Hz for BridgeData~V2~\cite{walke2023bridgedata}, $3$~Hz for Fractal~\cite{collaboration2023open}, and $20$~Hz for the RoboCasa-GR1~\cite{bjorck2025gr00t} and RoboCasa365~\cite{nasiriany2026robocasa365} corpora. EgoDex~\cite{hoque2025egodex} is the exception, subsampled by $3{\times}$ to $10$~Hz following H-RDT~\cite{bi2026h} and $\Psi_0$~\cite{wei2026psi_0}. The flow-matching target is a latent $z_h$ sampled from this posterior. We train with classifier-free guidance at an instruction-dropout rate of $0.1$.

\mypara{UniVLA VQ teacher.} The LAM teacher ablation in \tabref{tab:ablation} keeps the student, losses, and Stage~II recipe fixed and replaces only the teacher with the discrete LAM released by UniVLA~\cite{bu2025univla}, which encodes each frame transition into four tokens drawn from a $16{\times}128$ codebook. Evaluation follows the three-task WidowX protocol used in the ablations (\tabref{tab:ablation}).

\subsection{Pretraining Corpus}
Stage~I pretrains on a label-free mixture of $556{,}890$ trajectories spanning egocentric human video and four robot embodiments (\tabref{tab:sup_corpus}). It pairs a large human prior from EgoDex~\cite{hoque2025egodex} with single-arm manipulation from BridgeData~V2~\cite{walke2023bridgedata} and Fractal~\cite{collaboration2023open}, simulated GR1 humanoid manipulation from RoboCasa-GR1~\cite{bjorck2025gr00t}, real GR1 manipulation from DreamDojo-GR1~\cite{gao2026dreamdojo}, and simulated kitchen manipulation from RoboCasa365~\cite{nasiriany2026robocasa365}. Spanning human, simulated, and real sources exposes the latent to diverse dynamics during pretraining.

\begin{table*}[t]
    \centering
    \resizebox{0.99\linewidth}{!}{
    \begin{tabular}{y{30mm}|x{32mm}x{22mm}x{18mm}x{18mm}}
        \toprule
        \textbf{Pretraining corpus} & \textbf{Domain} & \textbf{Episodes} & \textbf{Frames} & \textbf{Rate (Hz)} \\
        \midrule
        EgoDex        & Human, egocentric    & $338{,}234$ & $90.0$M & $30{\to}10$ \\
        BridgeData~V2 & WidowX, real         & $53{,}192$  & $1.9$M  & $5$  \\
        Fractal       & Google robot, real   & $87{,}212$  & $3.8$M  & $3$  \\
        RoboCasa-GR1  & GR1, simulation      & $24{,}000$  & $6.0$M  & $20$ \\
        DreamDojo-GR1 & GR1, real            & $22{,}209$  & $6.4$M  & $20$ \\
        RoboCasa365   & RoboCasa, simulation & $32{,}043$  & $29.1$M & $20$ \\
        \midrule
        Total         &                      & $556{,}890$ & $137.2$M & \\
        \bottomrule
    \end{tabular}
}

    \caption{Stage~I pretraining corpus. Frames are native-rate totals, though we subsample only EgoDex to $10$~Hz.}
    \label{tab:sup_corpus}
\end{table*}

\subsection{Training and Inference Configuration}
Stage~I optimizes $\mathcal{L}_{\text{total}}$ on action-free videos for $250$k steps at batch size $128$, without any explicit action labels, weighting the distillation and motion terms by $\lambda_{\text{d}}{=}0.05$ and $\lambda_{\text{m}}{=}0.5$. We concatenate all corpora into one pool and sample uniformly from it, so each corpus contributes in proportion to its size. The flow-matching term regresses the velocity $v_\theta$ along the linear interpolant $x_{\tau,h}{=}(1-\tau)x_{0,h}+\tau z_h$ toward $v{=}z_h-x_{0,h}$, sampling $\tau\sim\mathcal{U}(0,1)$ under an MSE loss. The motion predictor supervises per-patch residuals at $M{=}4$ evenly strided steps ($h\in\{3,7,11,15\}$) of the $H{=}16$ chunk for training efficiency. Stage~II trains a fresh action decoder on the target embodiment's labeled actions and adapts the latent policy (\tabref{tab:sup_embodiments}). Both stages use AdamW with a cosine learning-rate schedule, bf16-mixed precision, gradient clipping at $1.0$, and weight EMA at decay $0.9999$. The learning rate warms up linearly over the first $5\%$ of steps to its peak (\tabref{tab:sup_hparams}), then decays to $0.1{\times}$ the peak in Stage~I and $0.01$--$0.05{\times}$ in Stage~II. In Stage~II, the latent chunk is produced by $4$-step Euler integration of the velocity field, and the action decoder maps it to executable commands.

\subsection{Per-Embodiment Configuration}
Stage~II attaches an $8$-block action decoder that translates the latent chunk to a $d_a$-dimensional action conditioned on a $d_p$-dimensional proprioceptive state, and adapts the shared latent policy at horizon $H{=}16$. For EgoDex, we freeze the latent policy because the DreamDojo teacher was pretrained on EgoDex and its latent already captures the domain, and train only the decoder. For the other embodiments, we finetune the latent policy at $1{\times}10^{-5}$, $0.1{\times}$ the decoder learning rate. We normalize every proprioception (and target action) dimension to $[-1,1]$ by $x \mapsto \mathrm{clip}\big(2(x-q_{01})/(q_{99}-q_{01})-1,\,-1,\,1\big)$, where $q_{01}$ and $q_{99}$ are the per-dimension percentiles of each corpus. GR1 first encodes its joint angles as $[\sin q, \cos q]$, which doubles its proprioception to $58$ dimensions, whereas the other embodiments normalize the raw state. \tabref{tab:sup_embodiments} reports the per-embodiment $d_a$, $d_p$, freeze setting, and Stage~II step budget, while \tabref{tab:sup_hparams} collects the optimization settings shared across embodiments.

\section{Evaluation Benchmarks} 
\label{sec:sup-benchmarks}

We evaluate \ours{} across four evaluation settings: egocentric human, real-to-sim single-arm robots, a humanoid, and a physical robot. \tabref{tab:sup_embodiments} summarizes each benchmark's embodiment, action and proprioception dimensions, task count, metric, and Stage~II adaptation. We detail EgoDex, SimplerEnv, and RoboCasa-GR1 below and cover the physical robot in \secref{sec:sup-realrobot}.

\begin{table*}[t]
    \centering
    \resizebox{\linewidth}{!}{
    \begin{tabular}{y{31mm}|y{26mm}|y{26mm}|x{4mm}|x{7mm}|x{24mm}|x{12mm}|x{8mm}}
        \toprule
        \textbf{Benchmark} & \textbf{Embodiment} & \textbf{Action space} ($d_a$) & $d_p$ & \textbf{Tasks} & \textbf{Metric} & \textbf{LAP} & \textbf{Steps} \\
        \midrule
        EgoDex            & Egocentric human  & $48$-D hand delta    & $48$ & --  & ADE\,/\,FDE ($m$) & frozen    & $40$k  \\
        SimplerEnv-WidowX & WidowX arm        & $7$-D EE (delta) & $8$  & $4$ & Success rate (\%) & finetuned & $100$k \\
        SimplerEnv-Google & Google robot arm  & $7$-D EE (delta) & $8$  & $3$ & Success rate (\%) & finetuned & $30$k  \\
        RoboCasa-GR1      & GR1 humanoid      & $29$-D joints       & $58$ & $6$ & Success rate (\%) & finetuned & $100$k \\
        Real world (ours) & xArm7 arm         & $8$-D joints        & $8$  & $3$ & Success rate (\%) & finetuned & $20$k  \\
        \bottomrule
    \end{tabular}
}

    \caption{Per-embodiment overview. The action space gives the action dimension $d_a$ and its form, $d_p$ is the proprioception dimension, EE denotes end-effector, LAP marks whether the latent policy is frozen or finetuned, and Steps is the Stage~II adaptation budget.}
    \label{tab:sup_embodiments}
\end{table*}

\mypara{EgoDex}~\cite{hoque2025egodex} is an egocentric dataset of human dexterous manipulation captured with Apple Vision Pro, with $338$k trajectories ($90$M frames) across $194$ tasks. The $48$-D action is the camera-frame delta of a bimanual hand pose, comprising a $9$-D wrist pose ($3$-D position and $6$-D rotation) and five fingertip positions ($5{\times}3$) per hand. We adopt the official train/test split and report on its $3{,}243$-trajectory test set, downsampling the $30$~Hz capture to $10$~Hz. From a single egocentric RGB frame and the current hand pose, \ours forecasts a $16$-step ($1.6$~s) chunk. We score Average and Final Displacement Error (ADE\,/\,FDE, $m$) in the camera frame over the $12$ keypoints (two wrists and ten fingertips), aggregated as best-of-$S$ over $S\in\{1,5,10\}$ following the official evaluation protocol.

\mypara{SimplerEnv}~\cite{li2024evaluating} provides reproducible real-to-sim proxies for single-arm manipulation, and we follow the evaluation setup of StarVLA~\cite{community2026starvla}. We train the action decoder on BridgeData~V2~\cite{walke2023bridgedata} for the \emph{WidowX} suite of four tabletop tasks (\emph{Put Spoon on Towel}, \emph{Put Carrot on Plate}, \emph{Stack Green Cube on Yellow Cube}, and \emph{Put Eggplant in Yellow Basket}), and on Fractal~\cite{collaboration2023open} for the \emph{Google Robot} suite of three tasks (\emph{Pick Coke Can}, \emph{Move Near}, and \emph{Open/Close Drawer}). We report per-task and average success rate, with $24$ trials per WidowX task and $300$, $240$, and $216$ trials for the Google Robot \emph{Pick Coke Can}, \emph{Move Near}, and \emph{Open/Close Drawer} tasks respectively. WidowX runs at $5$~Hz control with a $120$-step episode, and Google Robot at $3$~Hz with an $80$-step episode ($113$ for the drawer). We evaluate WidowX under Visual Matching, and Google Robot under both Visual Matching and Variant Aggregation, which perturbs background, lighting, distractor objects, and surface texture.

\mypara{RoboCasa-GR1}~\cite{bjorck2025gr00t} offers humanoid tabletop manipulation in the RoboCasa platform~\cite{nasiriany2026robocasa365}. The GR1 humanoid has two $7$-DoF arms, two $6$-DoF dexterous hands, and a $3$-DoF waist, giving a $29$-D joint-space action. The model observes a single egocentric RGB view. We evaluate $6$ articulated multi-step tasks. Each pairs an object with a cabinet, drawer, or microwave and issues an instruction such as ``pick up the bottle, place it into the cabinet, and close the cabinet''. Success requires shutting the receptacle's door, and \tabref{tab:gr1_detail} abbreviates the tasks as, e.g., \emph{BottleToCabinet}. We learn the action decoder from $1000$ demonstrations per task and report success over $50$ rollouts per task, evaluated in $10$ parallel environments at a $720$-step episode.

\section{Real-Robot Experiments} \label{sec:sup-realrobot}

\subsection{Setup}

\begin{figure}[!htbp]
    \centering
    \includegraphics[width=0.6\linewidth]{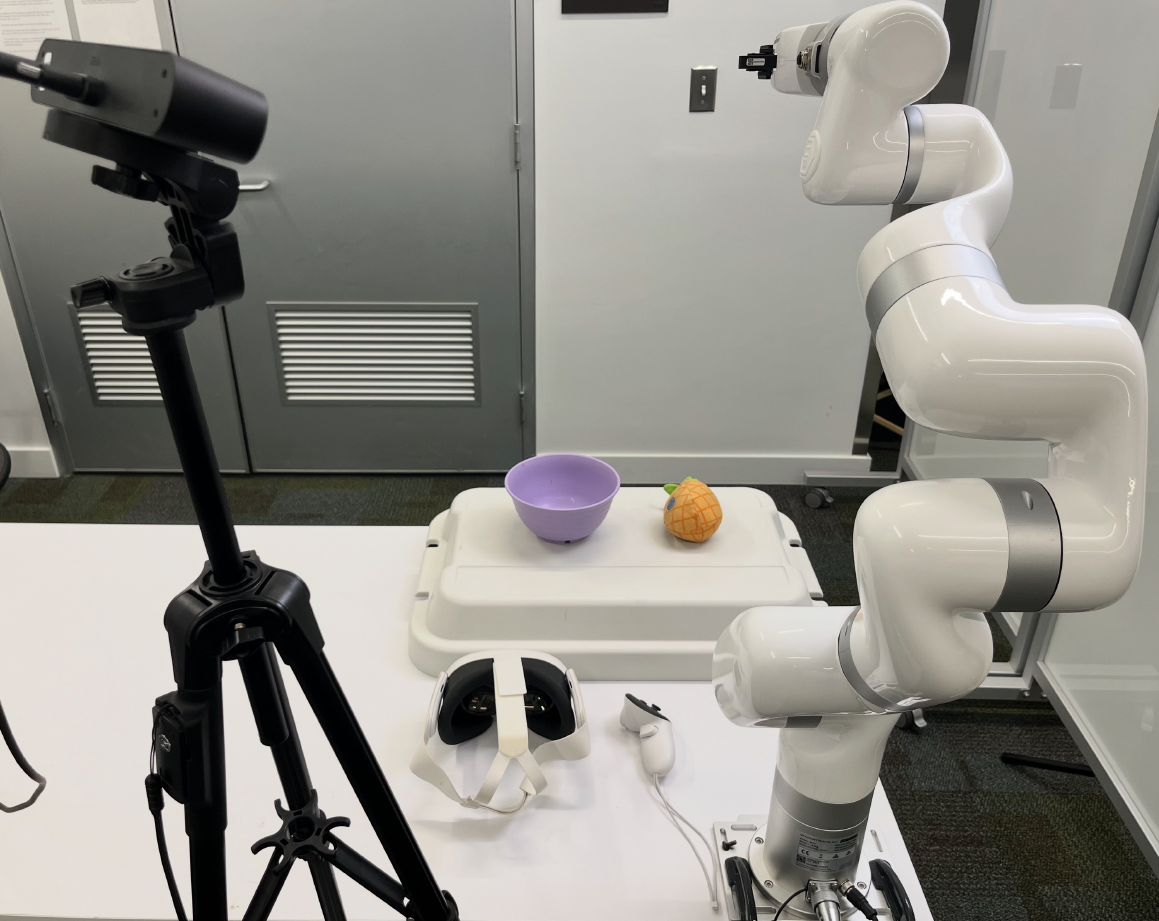}
    \caption{Real-robot platform: the $7$-DoF xArm7 with a parallel-jaw gripper, the tripod-mounted ZED~2i that supplies the fixed third-person view, and the Meta~Quest controllers used for VR teleoperation during data collection.}
    \label{fig:sup_platform}
\end{figure}

\mypara{Platform.} We use a $7$-DoF UFACTORY xArm7 with a parallel-jaw gripper and a single uncalibrated Stereolabs ZED~2i stereo camera mounted on a tripod at the left rear of the arm (\figref{fig:sup_platform}). We take its left-eye RGB at HD720 ($1280{\times}720$) as the fixed third-person view, and our model operates on a $224{\times}224$ input. The view feeds \ours directly, and no extrinsic calibration is needed.

\mypara{Observation and action.} At each step, \ours receives the current left-view RGB frame and an $8$-D proprioceptive vector of $7$ arm joint angles and the gripper opening. It predicts a chunk of $16$ actions, each $8$-D, comprising relative joint-position targets $q_{t+1+h}-q_t$ and an absolute normalized gripper command. At execution, predicted joint offsets are denormalized and added to the measured configuration.

\begin{figure}[!htbp]
    \centering
    \includegraphics[width=\linewidth]{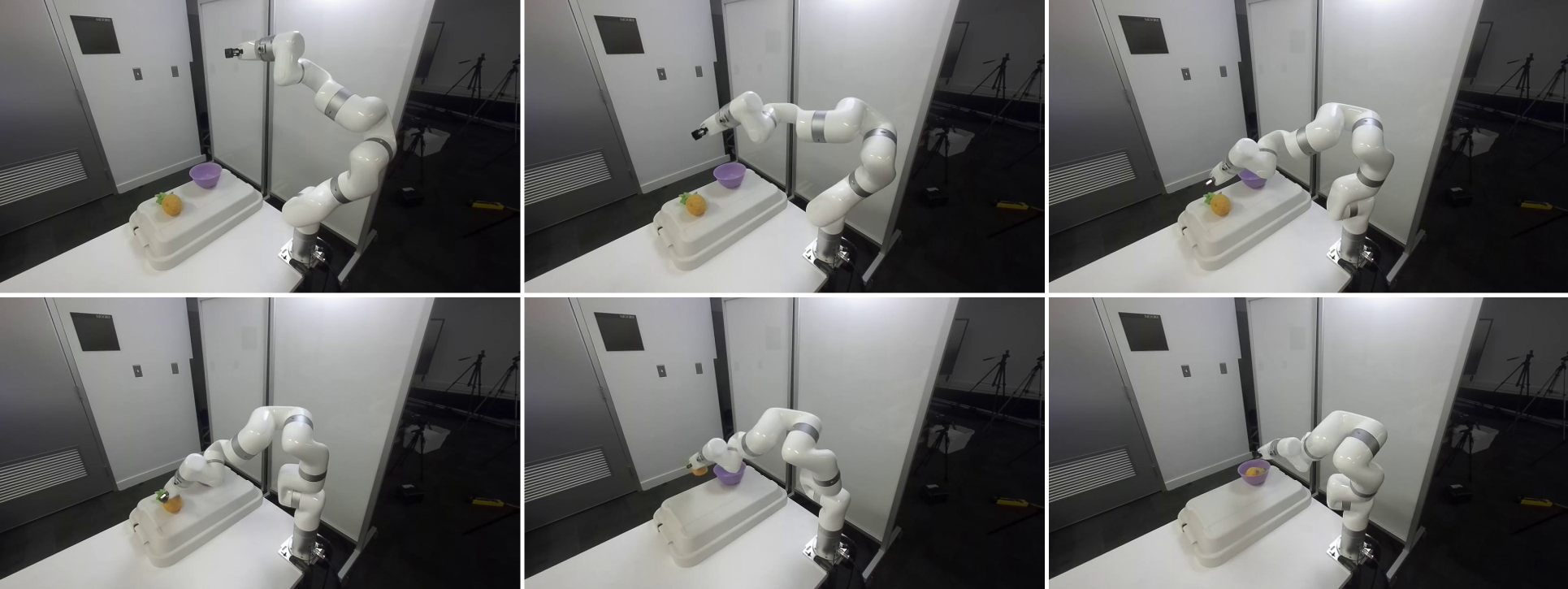}
    \caption{Teleoperated demonstration of \emph{T1: Pineapple$\rightarrow$Bowl} in the robot's observation view, read left to right, top to bottom. Zoom in for better viewing.}
    \label{fig:sup_demo_t1}
\end{figure}

\begin{figure}[!htbp]
    \centering
    \includegraphics[width=\linewidth]{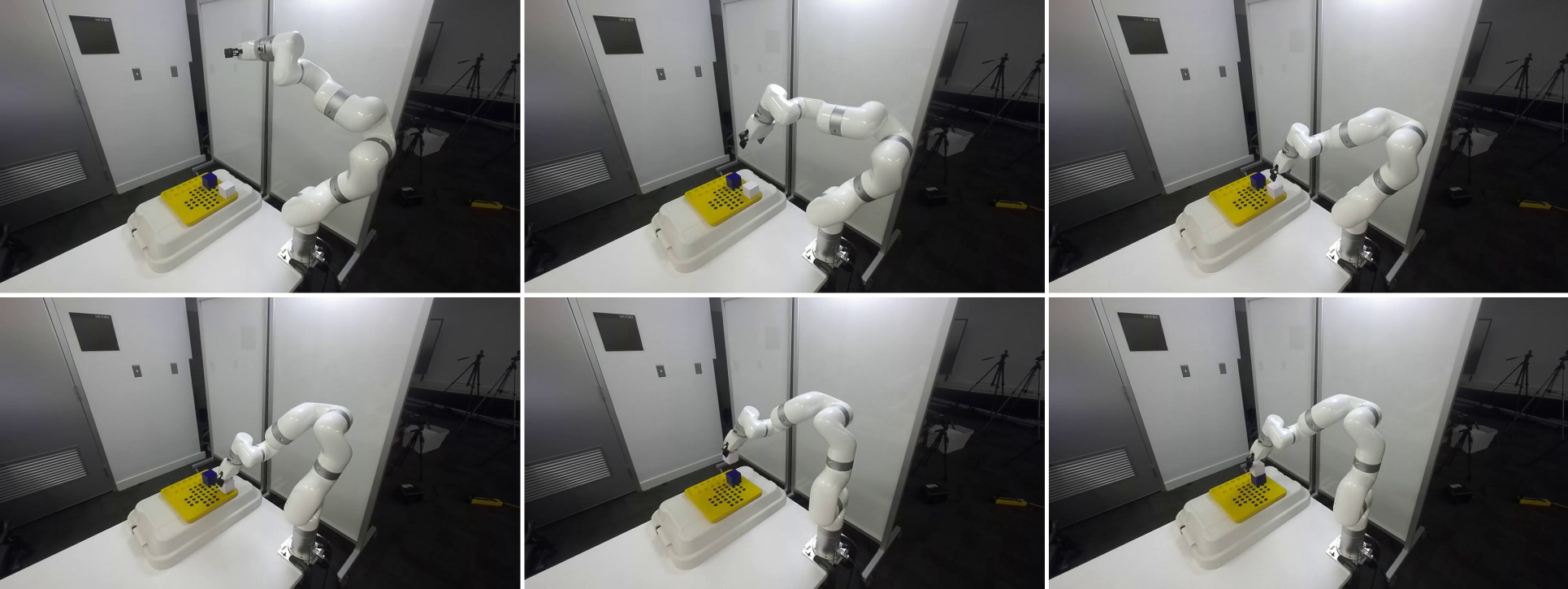}
    \caption{Teleoperated demonstration of \emph{T2: Stack Cubes} in the robot's observation view, read left to right, top to bottom. Zoom in for better viewing.}
    \label{fig:sup_demo_t2}
\end{figure}

\begin{figure}[!htbp]
    \centering
    \includegraphics[width=\linewidth]{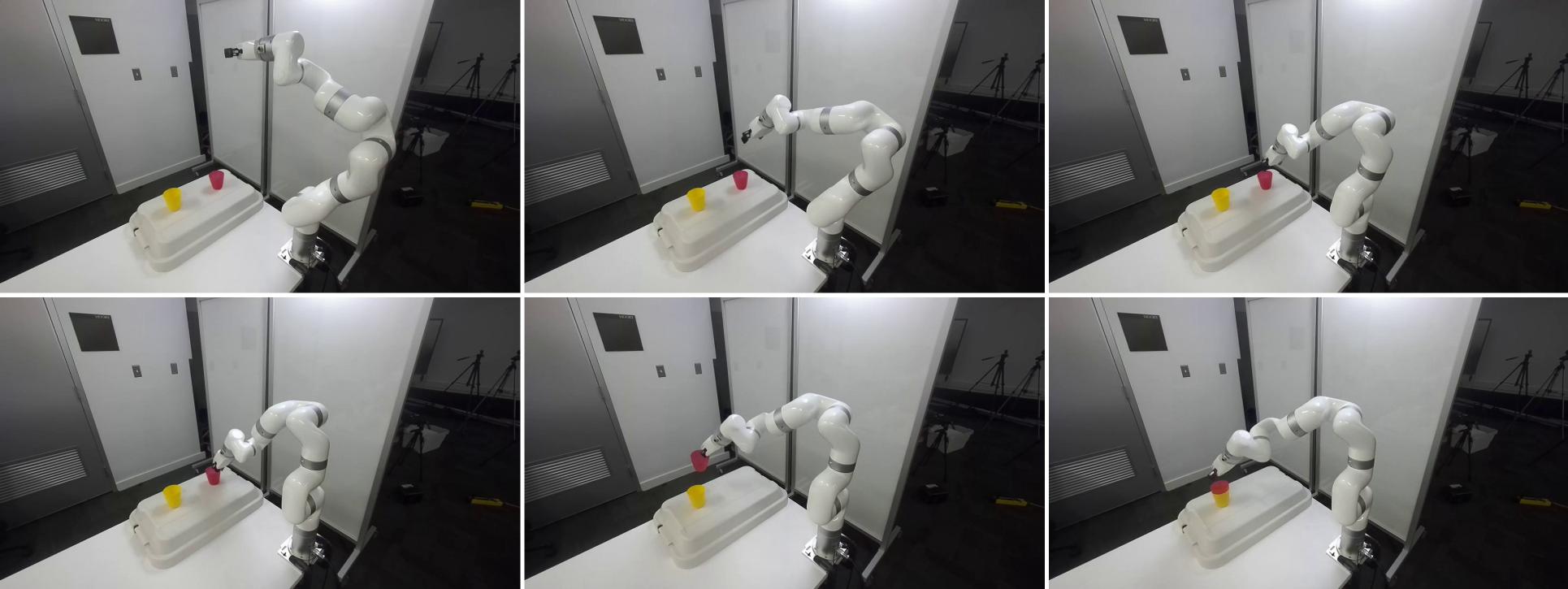}
    \caption{Teleoperated demonstration of \emph{T3: Stack Cups} in the robot's observation view, read left to right, top to bottom. Zoom in for better viewing.}
    \label{fig:sup_demo_t3}
\end{figure}

\mypara{Data collection.} We collect $50$ demonstrations per task ($150$ total) via Meta~Quest VR teleoperation that streams end-effector targets to the arm. \figref{fig:sup_demo_t1}-\ref{fig:sup_demo_t3} show one teleoperated demonstration snapshot for each task. Proprioception and the teleoperation command stream are logged at $100$~Hz alongside the ZED stereo recording.

\mypara{Data preprocessing.} Each demonstration is uniformly subsampled in time to a fixed length of $49$ frames. For each target timestamp we take the temporally nearest left-eye RGB frame, with alignment error within one camera frame (${\le}\,44$~ms), together with the robot state and action sampled from the $100$~Hz log, giving a one-to-one correspondence between frames and action targets. Clips are packaged in the LeRobot~v2.0 format at a nominal $16$~fps. Because every variable-duration episode is normalized to $49$ frames, the effective temporal stride varies across demonstrations.

\mypara{Deployment and control rates.} The policy runs as an inference server that a robot-side client queries over ZMQ. Action chunking decouples the policy from the low-level controller across three rates, described next from fastest to slowest. The \emph{servo rate} ($200$~Hz) is the rate at which the client sends a joint target to the arm's low-level controller, one every $5$~ms, and is independent of model inference. The \emph{action rate} ($4.8$~Hz) is the rate at which the chunk's $16$ steps execute, the $16$~fps demonstration frame rate scaled by a $0.3{\times}$ deployment speed, with each step interpolated into ${\approx}\,42$ servo commands. The \emph{replanning rate} (${\sim}\,0.28$~Hz) is the rate at which the client queries a fresh chunk after executing all $16$ steps. Each cycle takes ${\approx}\,3.5$~s, including action execution and the next query, while the arm tracks each chunk open-loop. One query takes ${\sim}\,188$~ms end to end (median) on an RTX~PRO~5000 Laptop GPU, covering serialization, transport, and the model forward. Chunking lets the servo loop replay the precomputed chunk while no query is in flight, so this latency leaves the $200$~Hz servo rate intact. It does age each chunk's observation by that amount and stall the loop once per chunk, lowering the achieved action rate from $4.8$ to ${\sim}\,4.5$~Hz.

\subsection{Task Descriptions}
The three evaluated tabletop tasks share a pick-and-place structure but escalate in the precision of the placement sub-skill. We additionally probe compositional generalization with an out-of-distribution task.

\mypara{T1: Pineapple$\rightarrow$Bowl.} A tolerant drop-in placement with an irregularly shaped object, where a coarse release above the bowl suffices. The instruction is \texttt{pick up the pineapple and put it into the bowl}.

\mypara{T2: Stack Cubes.} Stacking a white cube on a blue cube, which requires a precise grasp and aligned placement against soft-contact rebound from the compliant foam. We add a yellow background board, in both data collection and deployment, so the white cube is distinguishable from the white background. The instruction is \texttt{stack the white cube on the blue cube}.

\mypara{T3: Stack Cups.} Nesting a red cup on a yellow cup, which demands accurate alignment within a small placement tolerance on thin-walled objects. The instruction is \texttt{stack the red cup on the yellow cup}.

\mypara{OOD: Cup$\rightarrow$Bowl.} An out-of-distribution compositional task that pairs the red cup (seen only in T3) with the bowl (seen only in T1) in a combination absent from training. The instruction is \texttt{pick up the red cup and put it into the bowl}.

\subsection{Qualitative Rollouts}
\begin{figure}[!htbp]
    \centering
    \includegraphics[width=\linewidth]{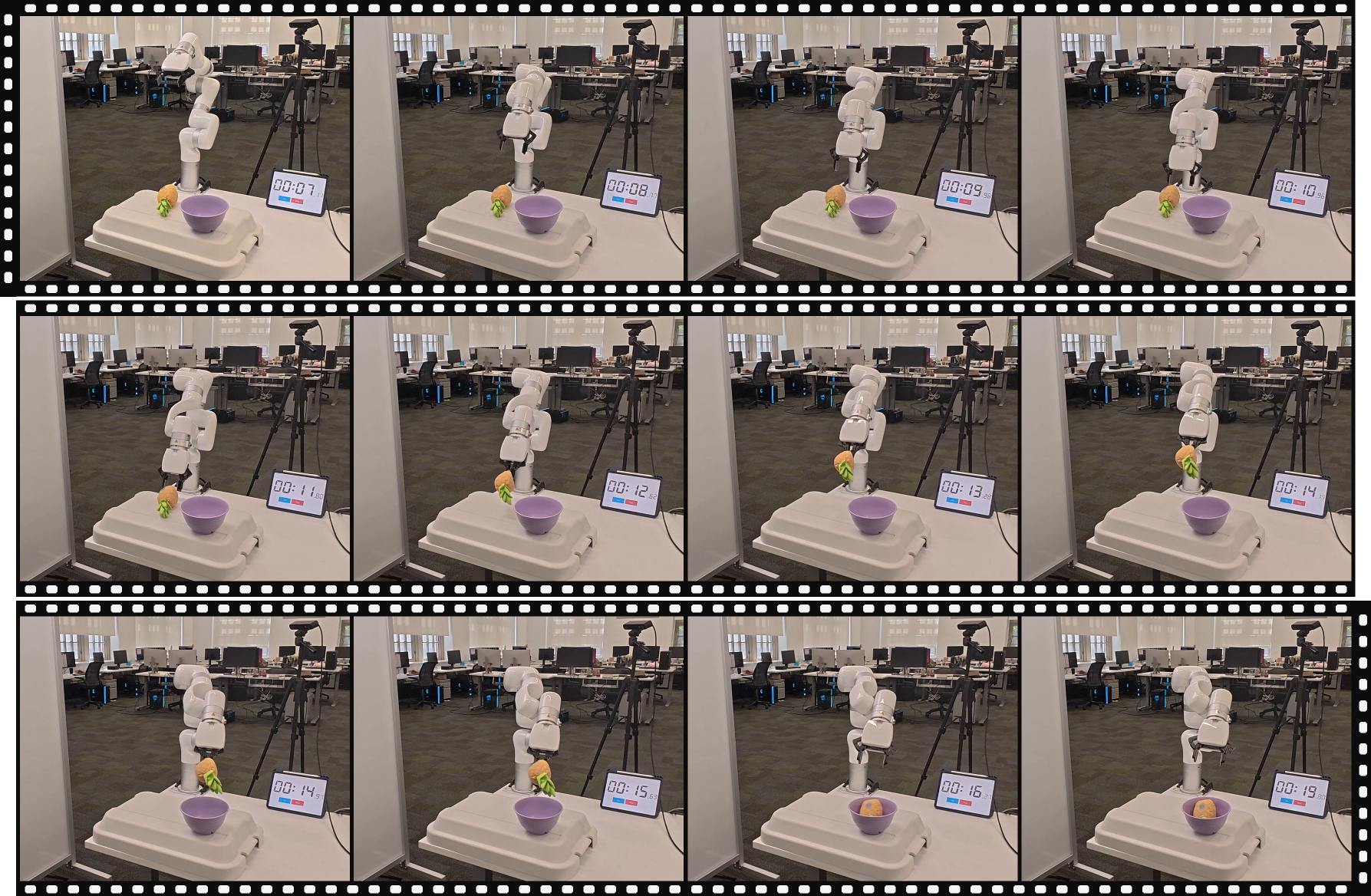}
    \caption{\ours{} rollout on \emph{T1: Pineapple$\rightarrow$Bowl}, key frames read left to right, top to bottom. See the supplementary video for the full rollout.}
    \label{fig:sup_rollout_pineapple}
\end{figure}

\begin{figure}[!htbp]
    \centering
    \includegraphics[width=\linewidth]{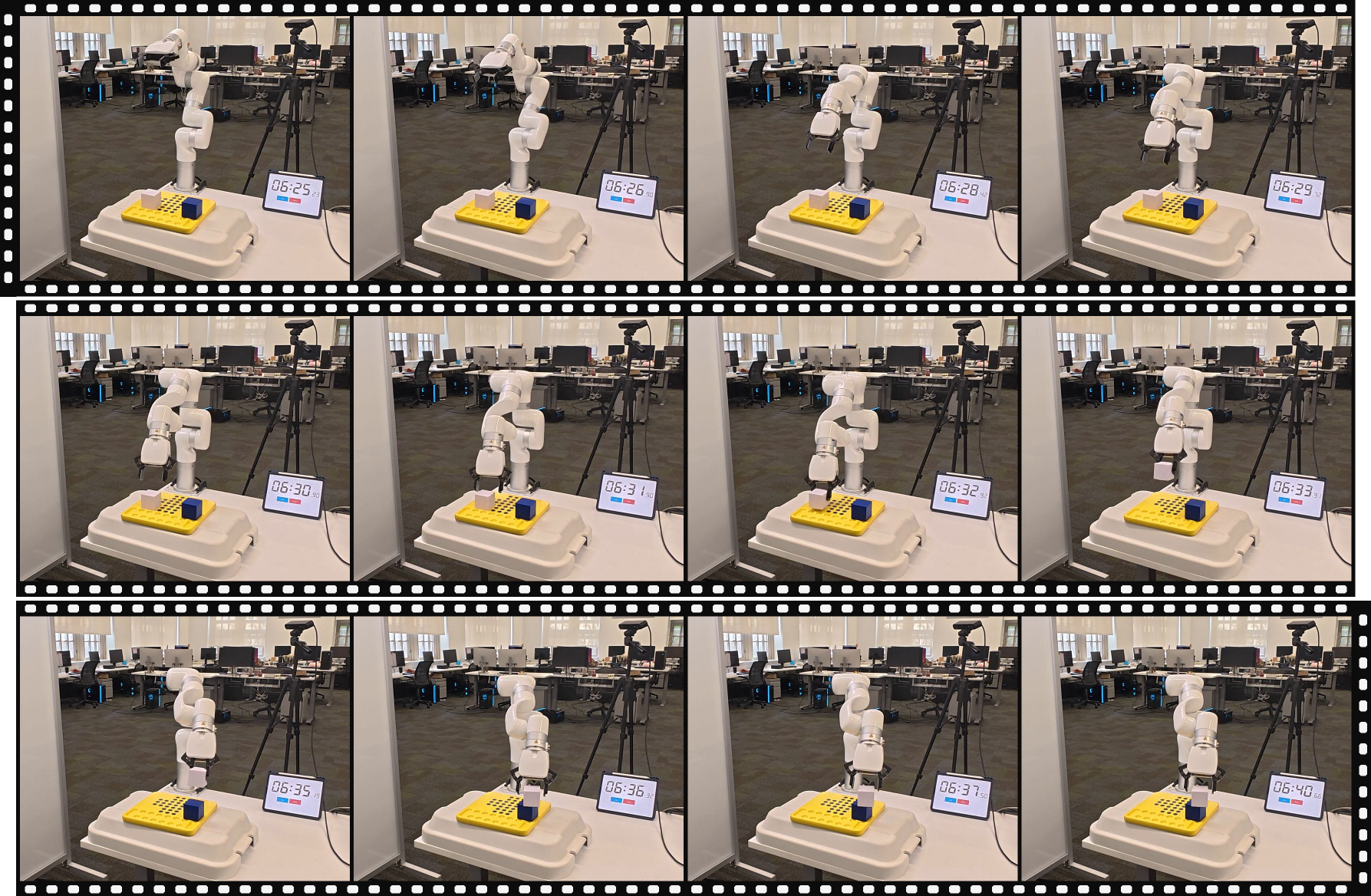}
    \caption{\ours{} rollout on \emph{T2: Stack Cubes}, key frames read left to right, top to bottom. See the supplementary video for the full rollout.}
    \label{fig:sup_rollout_cubes}
\end{figure}

\begin{figure}[!htbp]
    \centering
    \includegraphics[width=\linewidth]{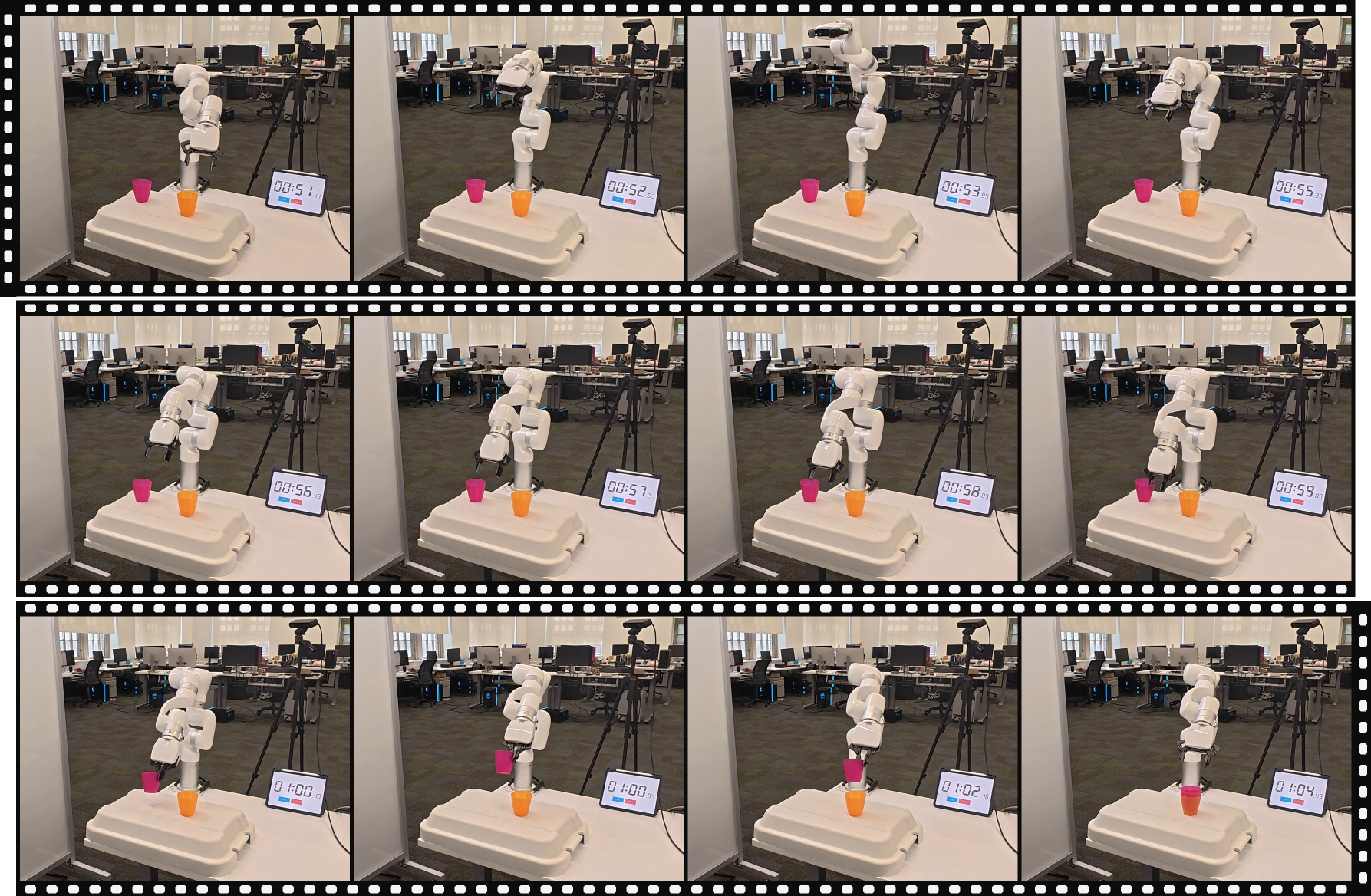}
    \caption{\ours{} rollout on \emph{T3: Stack Cups}, key frames read left to right, top to bottom. See the supplementary video for the full rollout.}
    \label{fig:sup_rollout_cups}
\end{figure}

\begin{figure}[!htbp]
    \centering
    \includegraphics[width=\linewidth]{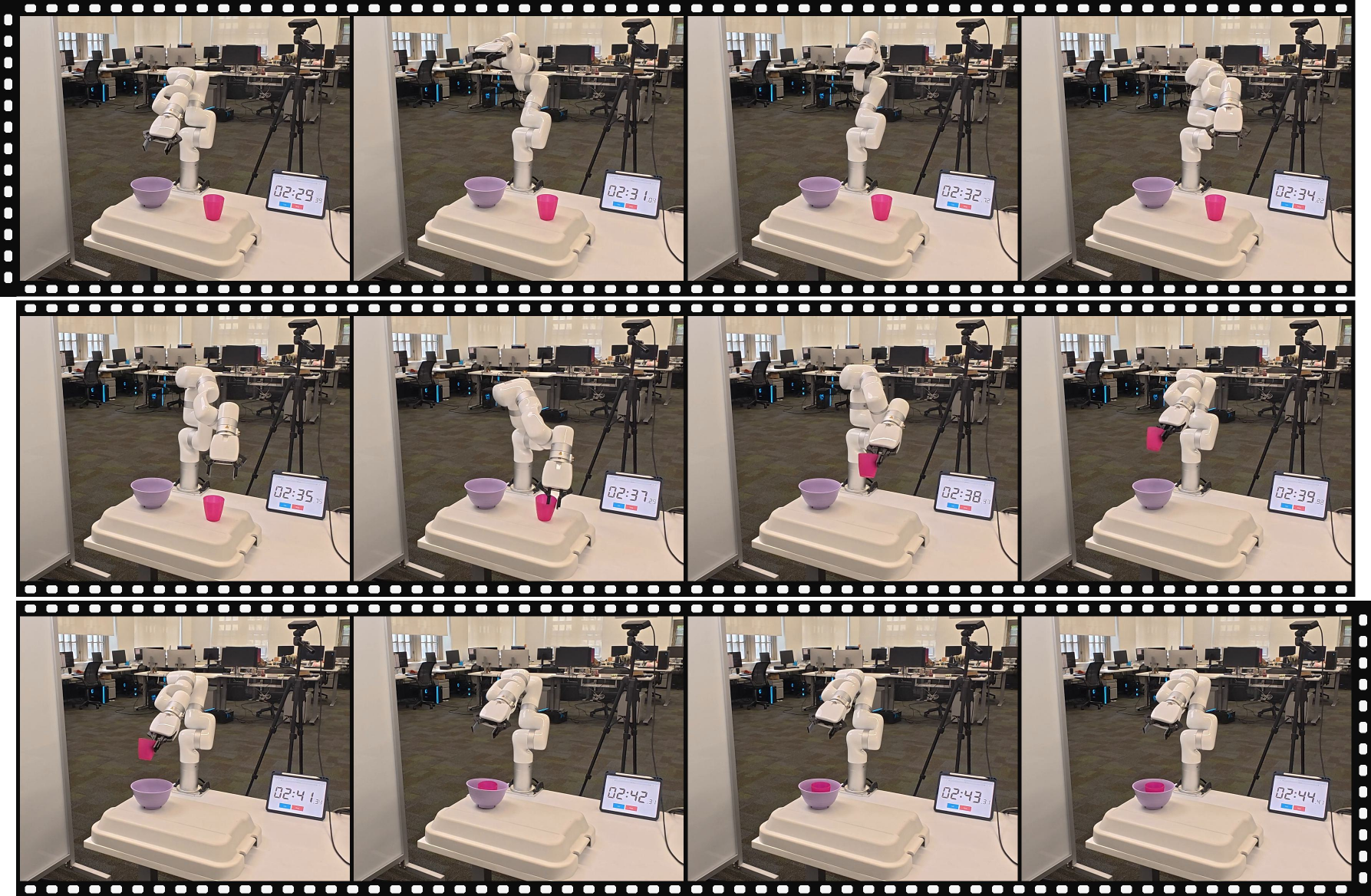}
    \caption{\ours{} rollout on the unseen compositional instruction \texttt{pick up the red cup and put it into the bowl}, key frames read left to right, top to bottom. See the supplementary video for the full rollout.}
    \label{fig:sup_rollout_ood}
\end{figure}

\figref{fig:sup_rollout_pineapple}-\ref{fig:sup_rollout_ood} show key-frame filmstrips of representative \ours{} rollouts on the three tabletop tasks and the unseen compositional instruction. We encourage viewing the supplementary video for the full sequences.

\subsection{Results and Per-Task Analysis}
Across $50$ rollouts per task, \ours{} averages $28.0\%$ against $\pi_{0.5}$'s $22.7\%$ (\tabref{tab:sup_realrobot}). The per-task rates track the precision ordering of the tasks and localize where the latent helps.

\begin{table*}[t]
    \centering
    \resizebox{0.8\linewidth}{!}{
    \begin{tabular}{y{26mm}|x{30mm}x{18mm}x{18mm}|x{12mm}}
        \toprule
        \textbf{Method} & Pineapple $\rightarrow$ Bowl & Stack Cubes & Stack Cups & \textbf{Average} \\
        \midrule
        $\pi_{0.5}$                                 & $42$         & $\textbf{12}$ & $14$          & $22.7$ \\
        \cmidrule(lr){1-5}
        \rowcolor{ourscolor} \textbf{\ours{(Ours)}} & $\textbf{46}$ & $10$          & $\textbf{28}$ & $\textbf{28.0}$ \\
        \bottomrule
    \end{tabular}
}

    \caption{Per-task real-robot success rate (\%) over $50$ rollouts, comparing against $\pi_{0.5}$~\cite{intelligence2025pi_}.}
    \label{tab:sup_realrobot}
\end{table*}

\mypara{Cube stacking.} Both $\pi_{0.5}$ and \ours{} collapse on cube stacking ($12\%$ and $10\%$), which marks a task-intrinsic difficulty. Two factors compound. First, the cube demonstrations show the largest cross-demonstration variability of the release moment and the highest gripper-channel prediction error, which points to the release sub-skill as the likely bottleneck. Second, the foam cubes are compliant, so a slightly mistimed release lets the top cube deform on contact and spring off the stack.

\mypara{Cup stacking.} Nesting one thin-walled cup on another demands accurate alignment within a small placement tolerance. The cup demonstrations are the most consistent in proprioception, with the smallest spread in episode length, gripper cycles, and release height, so the supervision is reliable and the difficulty lies in placement precision itself. Here \ours{} reaches $28\%$ and doubles $\pi_{0.5}$ at $14\%$, consistent with a continuous latent that preserves the fine motor distinctions.

\mypara{Compositional generalization.} An unseen instruction (``pick up the red cup and put it into the bowl'') pairs an object and a target drawn from different training tasks. \ours{} succeeds on $23$ of $50$ rollouts of this unseen composition, recombining learned grounding and skills rather than replaying a memorized trajectory.

\section{Additional Analysis} \label{sec:sup-analysis}

\subsection{Latent Probing} \label{sec:sup-probe}
We probe the latent's action content by fitting a linear regressor from each latent source to the ground-truth action chunk and reporting the held-out $R^2$. Evaluation windows are drawn from EgoDex, BridgeData~V2 (WidowX), and Fractal (Google Robot), about $3$k windows per corpus. The regression target is the recorded action of that corpus, namely the $48$-D camera-frame hand delta for EgoDex and the $7$-D end-effector delta for Bridge and Fractal. We compare three sources. The policy latent $\hat{Z}$ is the mean of $4$ flow-matching samples at guidance scale $1.0$. The teacher mean $\mu$ is the LAM posterior mean, which observes $o_{t+1}$ and acts as a non-causal reference. The appearance floor is a random projection of the frozen DINOv2 patch features to the latent dimension. Each probe is a ridge regressor whose penalty is selected by RidgeCV, and we report $R^2$ as the mean and standard deviation over $5$ held-out splits. For a latent source $x\in\mathbb{R}^{H d_z}$ and the flattened ground-truth action chunk $a\in\mathbb{R}^{C}$ with $C{=}H\,d_a$ channels, the ridge probe predicts $\hat{a}{=}Wx{+}b$, and we average the per-channel coefficient of determination over the held-out windows $n$,
\begin{equation}
    R^2 = \frac{1}{C}\sum_{c=1}^{C}\left(1 - \frac{\sum_{n}(a_{n,c}-\hat{a}_{n,c})^2}{\sum_{n}(a_{n,c}-\bar{a}_{c})^2}\right),
\end{equation}
with $\bar{a}_c$ the per-channel mean over those windows. The \emph{Recover} column reports the policy latent's score relative to the oracle teacher mean,
\begin{equation}
    \mathrm{Recover} = \frac{R^2(\hat{Z})}{R^2(\mu)},
\end{equation}
averaged over the splits where $R^2(\mu){>}0$.

The policy latent recovers $62.4\%$ of the oracle's $R^2$ on EgoDex, $56.5\%$ on Bridge, and matches it on Fractal, far above the appearance floor (\tabref{tab:probe}). The latent thus encodes action-predictive information from $o_t$ alone, not generic feature variance. 

\begin{table*}[t]
    \centering
    \resizebox{0.8\linewidth}{!}{
    \begin{tabular}{y{30mm}|x{20mm}x{20mm}x{20mm}}
        \toprule
        \multirow{2}{*}{\textbf{Latent source}} & \multicolumn{3}{c}{\textbf{Action} $R^2$ $\uparrow$} \\
        \cmidrule(lr){2-4}
            & EgoDex & Bridge & Fractal \\
        \midrule
        Random Proj.\ $f_t$             & $0.123_{\pm0.002}$ & $0.024_{\pm0.008}$ & $0.104_{\pm0.007}$ \\
        Teacher $\mu$ (oracle)      & $0.282_{\pm0.002}$ & $0.203_{\pm0.002}$ & $0.186_{\pm0.007}$ \\
        \cmidrule(lr){1-4}
        \rowcolor{ourscolor} \textbf{\ours{(Ours)}} & $0.176_{\pm0.003}$ & $0.115_{\pm0.006}$ & $0.194_{\pm0.009}$ \\
        \midrule
        \textbf{Recover}  & $62.4_{\pm0.5}\%$ & $56.5_{\pm3.3}\%$ & $104.1_{\pm3.0}\%$ \\
        \bottomrule
    \end{tabular}
}

    \caption{Aggregate linear probing of $\hat{Z}$ for action prediction on EgoDex, BridgeData~V2 (WidowX), and Fractal (Google Robot). Recover gives the policy latent's $R^2$ as a fraction of the oracle's.}
    \label{tab:probe}
\end{table*}

\subsection{Latent Multimodality} \label{sec:sup-multimodal}
We evaluate whether the policy's latent carries more than one plausible action for the same observation. For each of $512$ held-out EgoDex observations we draw $64$ latents and decode each one with the same frozen action decoder, so the decoding is deterministic and all spread in the terminal $12$-keypoint configuration comes from the latent. Standard unimodality tests are one dimensional, so we instead cluster the endpoints with $2$-means in their top three principal components and compare the resulting separation against $199$ draws from a Gaussian matched to the observed variance, rejecting at $\alpha = 0.05$. The null is rejected for $238$ of $512$ observations ($46.5\%$), and for $2$ of $100$ matched controls built from unimodal samples ($1$ of $100$ under a second seed). Among the rejected observations the centroid gap, measured between the two cluster centroids in the original coordinates, has median $4.53$\,cm with quartiles $2.98$ and $7.29$\,cm, comparable to the $S{=}1$ FDE of $4.7$\,cm, and the minority cluster holds a quarter of the samples. Non-rejection is not acceptance, so the remaining observations are labeled inconclusive in \figref{fig:multimodal} rather than unimodal.

\subsection{Language Sensitivity} \label{sec:sup-language}
We quantify the instruction sensitivity of the predicted latent (\figref{fig:motion}, \emph{left}). Holding the observation $o_t$ fixed, we perturb the instruction across five categories: paraphrase, object swap, verb change, location change, and full removal. For each, we compute the cosine distance between the predicted latent chunk $\hat{Z}$ under the original and perturbed instruction at a shared flow-matching seed, which isolates the instruction-induced change from sampling noise. We average this distance over $32$ observation frames and $4$ seeds. The figure shows raw cosine distances, while relative changes are measured against the paraphrase-level distance. Paraphrase and object swap remain at this floor, whereas verb change, location change, and removal induce progressively larger shifts, indicating that the latent encodes the target action and location rather than surface phrasing.

\subsection{Additional Hand Pose Forecasts} \label{sec:sup-egodex}
\begin{figure}[!htbp]
    \centering
    \includegraphics[width=\linewidth]{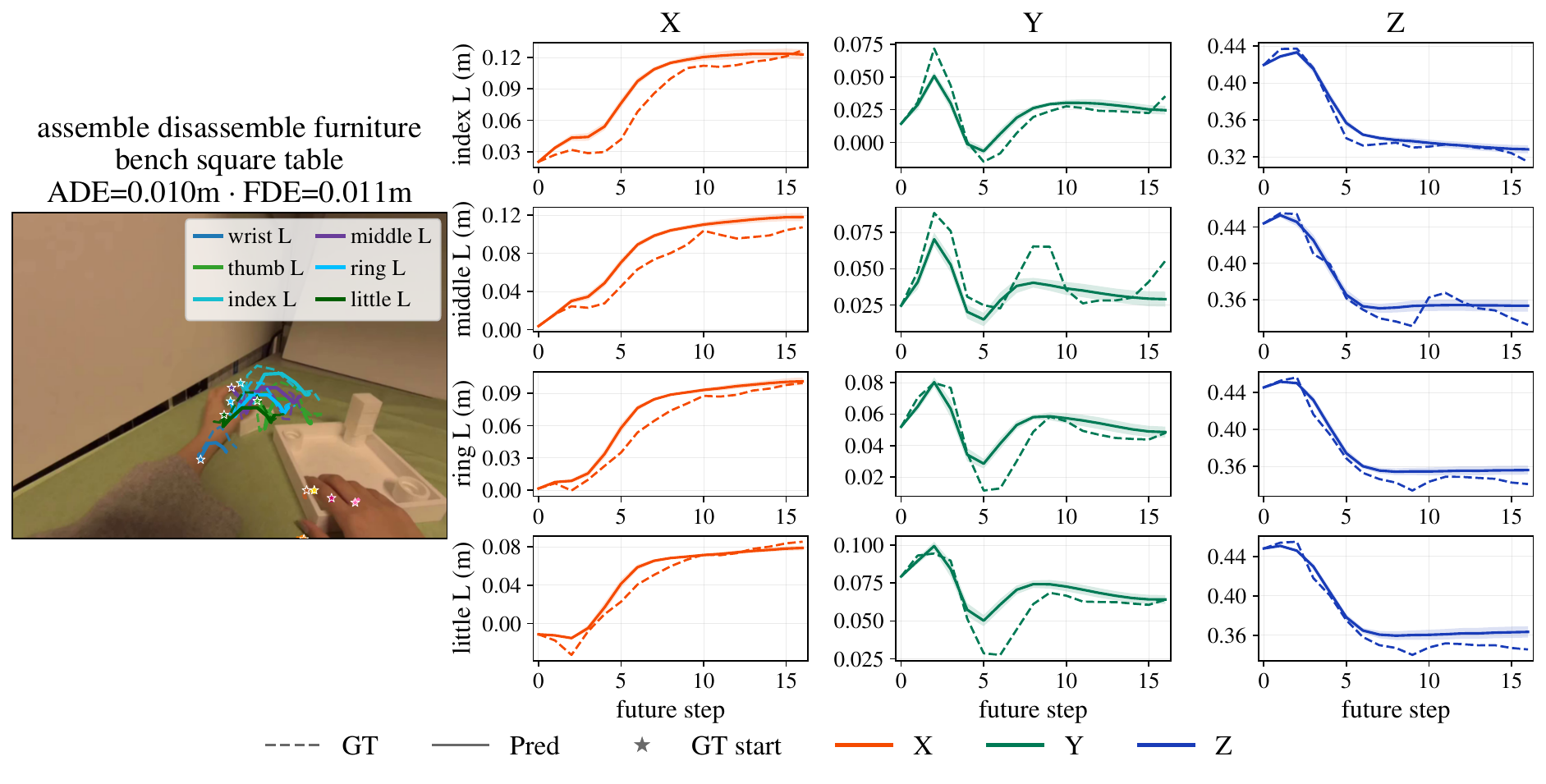}
    \caption{Four finger keypoints (index, middle, ring, little) of the acting hand for the \emph{assemble disassemble furniture} case of \figref{fig:egodex_traj}, which plots only the wrist and thumb. Each panel shows one keypoint's $X$/$Y$/$Z$ over the $1.6$\,s horizon, with ground truth dashed and prediction solid.}
    \label{fig:sup_furniture_fingers}
\end{figure}

\begin{figure}[!htbp]
    \centering
    \includegraphics[width=0.9\linewidth]{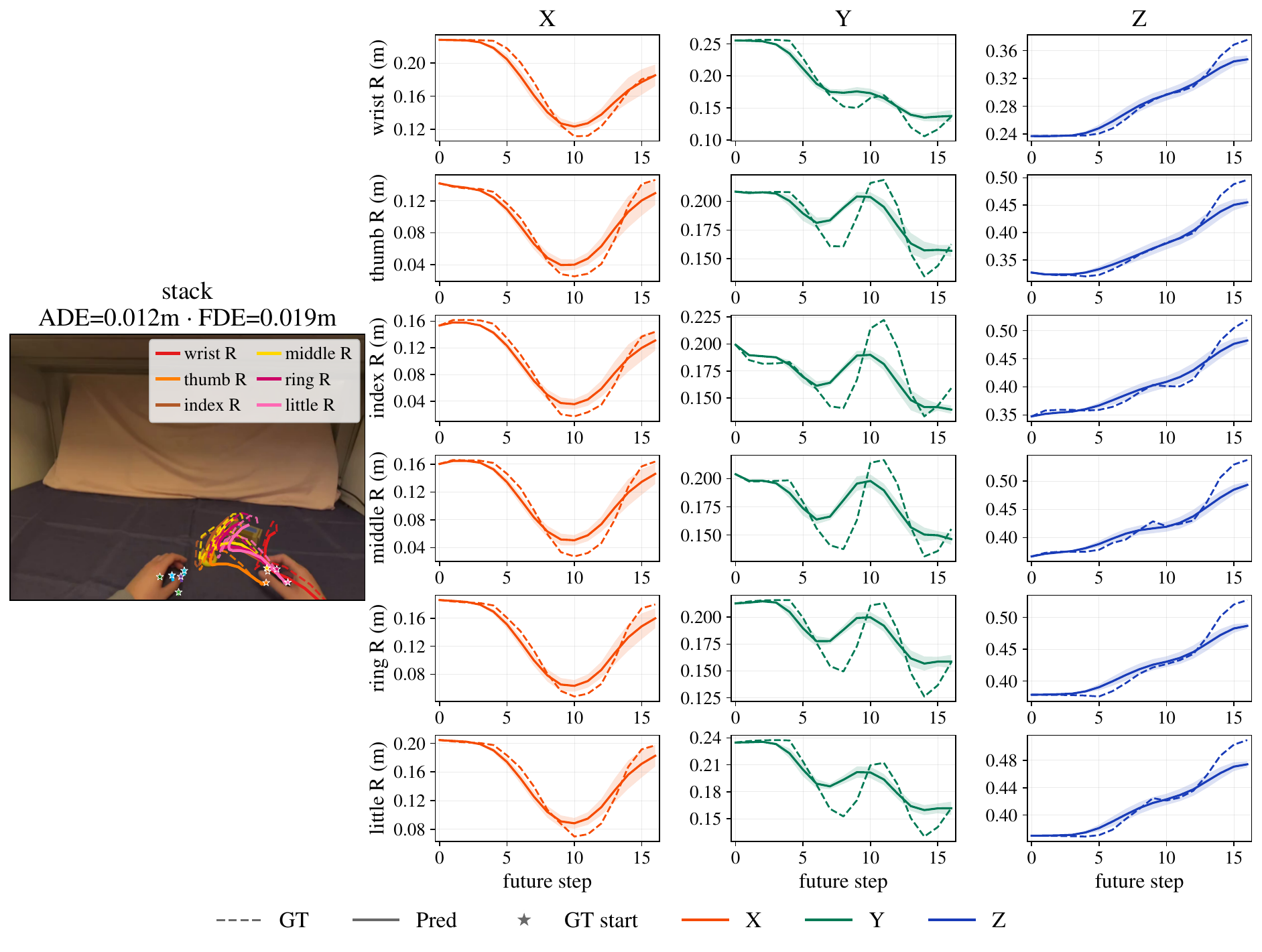}
    \caption{Full per-keypoint forecast on a \emph{stack} task, across all six keypoints of the acting hand. Panels follow \figref{fig:sup_furniture_fingers}: each shows one keypoint's $X$/$Y$/$Z$ over the $1.6$\,s horizon, with ground truth dashed and prediction solid.}
    \label{fig:sup_stack_full}
\end{figure}

\begin{figure}[!htbp]
    \centering
    \includegraphics[width=0.9\linewidth]{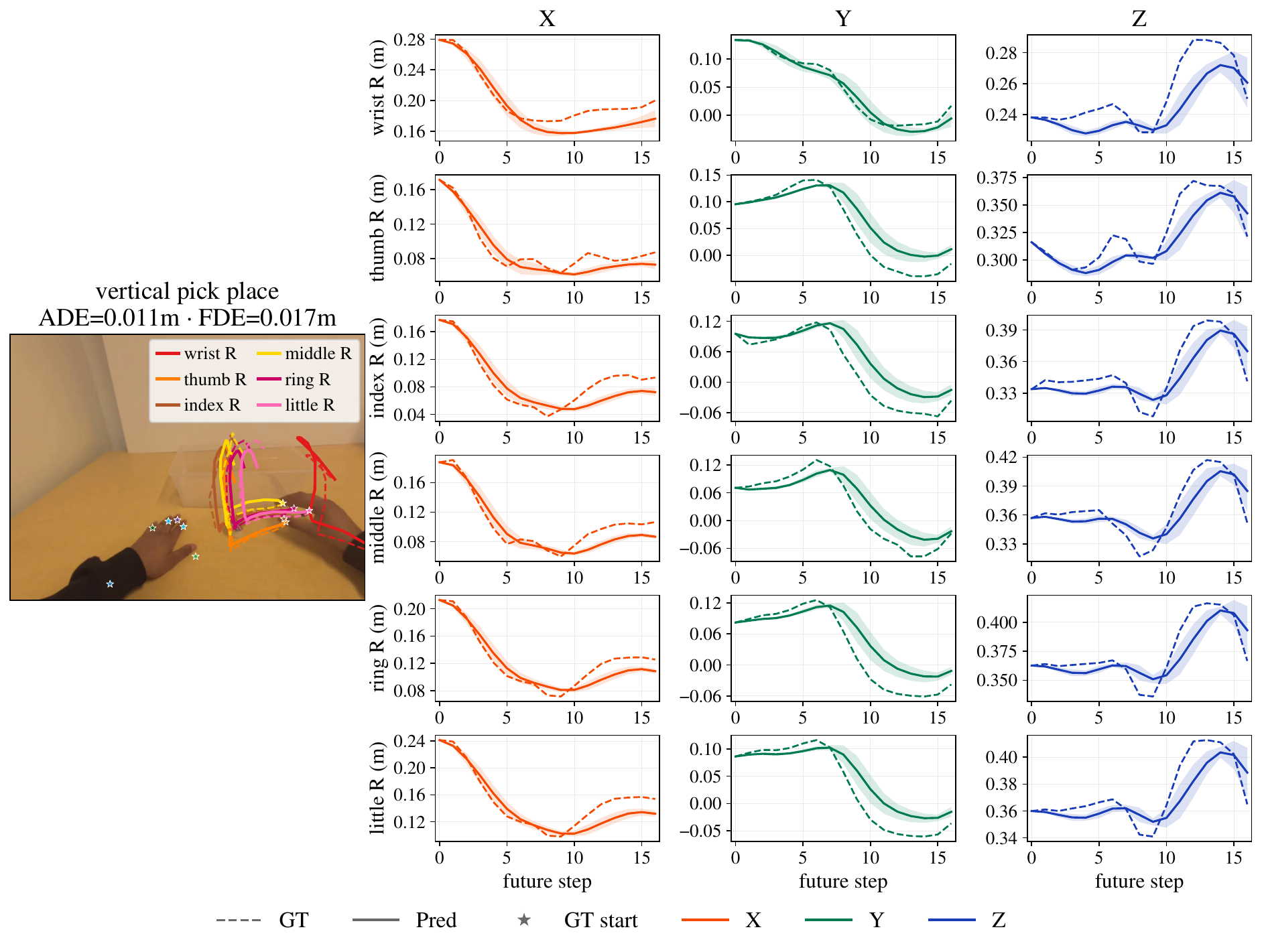}
    \caption{Full per-keypoint forecast on a \emph{vertical pick-and-place} scene, across all six keypoints of the acting hand. Panels follow \figref{fig:sup_furniture_fingers}: each shows one keypoint's $X$/$Y$/$Z$ over the $1.6$\,s horizon, with ground truth dashed and prediction solid.}
    \label{fig:sup_vpp_full}
\end{figure}

We extend the EgoDex qualitative study of \figref{fig:egodex_traj}, which plots the wrist and thumb of an assemble/disassemble furniture case. \figref{fig:sup_furniture_fingers} adds the four finger keypoints omitted from that case, and \figref{fig:sup_stack_full} and \figref{fig:sup_vpp_full} show the full six-keypoint set for a stack task and a vertical pick-and-place scene. The predictions broadly follow the ground-truth motion across keypoints, with local deviations over the forecasting horizon.

\subsection{Additional Latent Reconstructions} \label{sec:sup-recon}
\begin{figure}[!htbp]
    \centering
    \includegraphics[width=\linewidth]{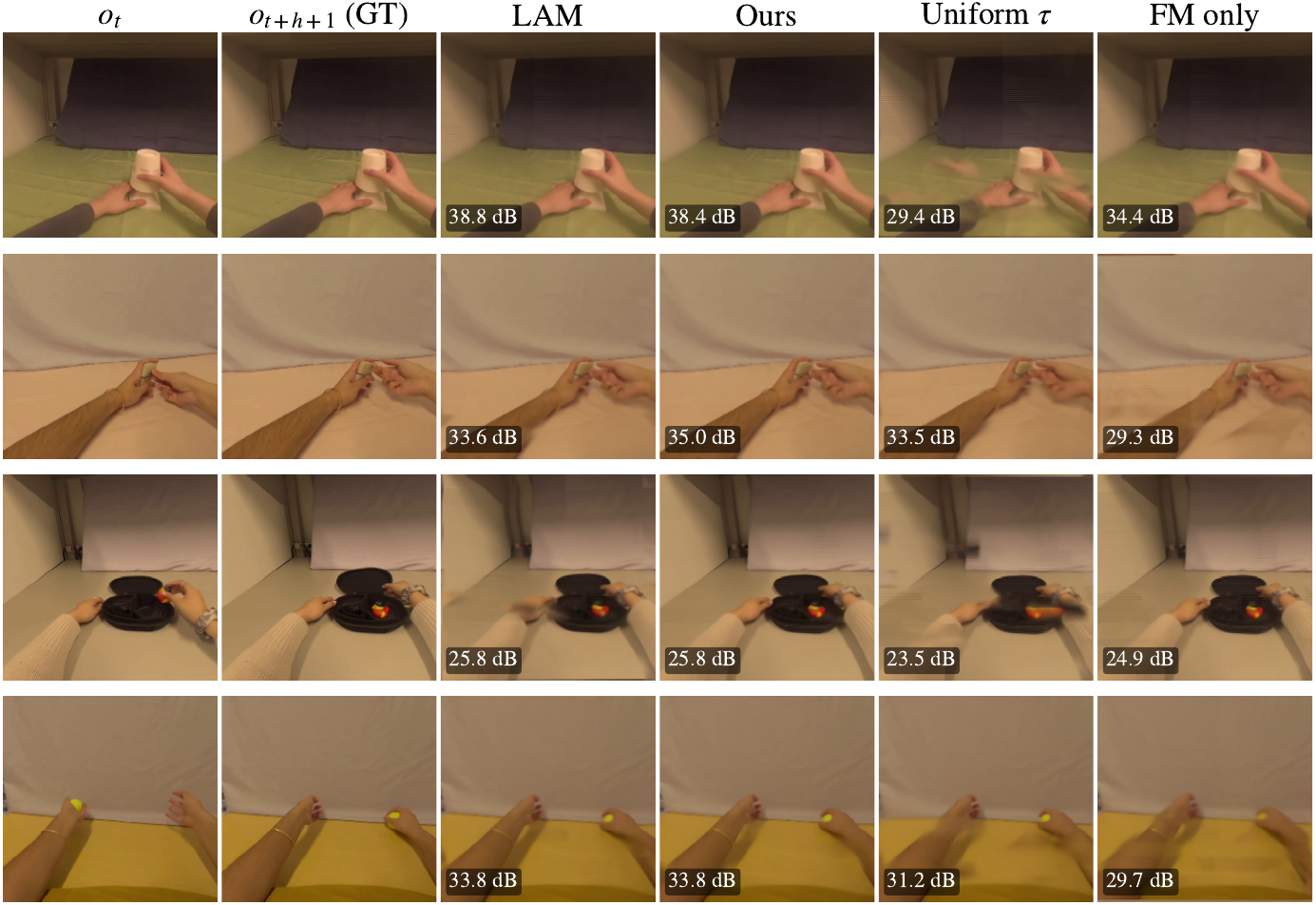}
    \caption{Additional latent reconstructions on EgoDex, four examples. Columns are $o_t$, the GT $o_{t+h+1}$, and reconstructions decoded from the teacher, \ours{}, Uniform~$\tau$, and FM-only latents, with PSNR against the GT. Each reconstruction is a single-step extrapolation from $o_{t+h}$, which is not shown; $o_t$ is given for reference.}
    \label{fig:sup_recon_egodex}
\end{figure}

\begin{figure}[!htbp]
    \centering
    \includegraphics[width=\linewidth]{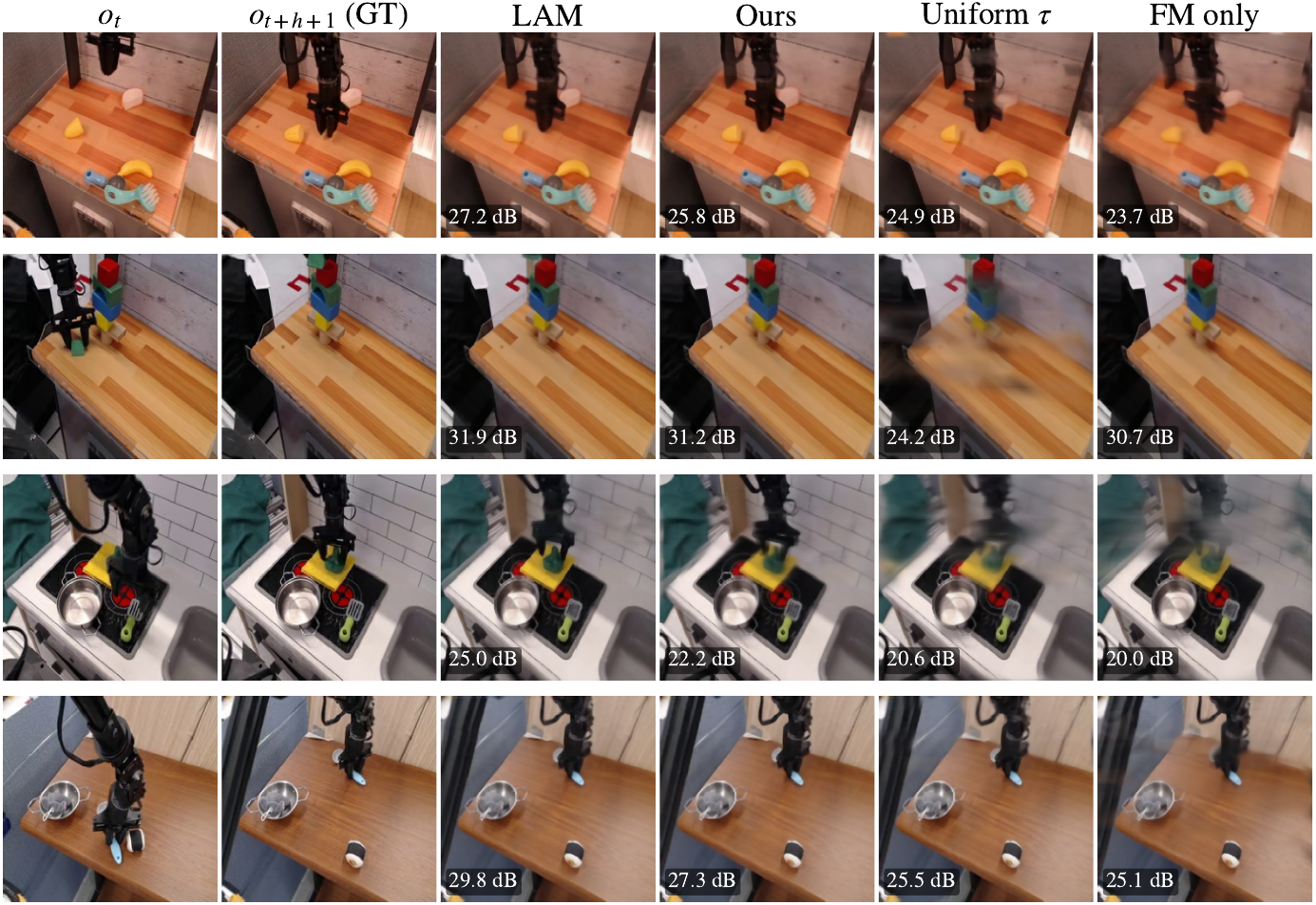}
    \caption{Additional latent reconstructions on WidowX (BridgeData~V2). Columns follow \figref{fig:sup_recon_egodex}.}
    \label{fig:sup_recon_bridge}
\end{figure}

\begin{figure}[!htbp]
    \centering
    \includegraphics[width=\linewidth]{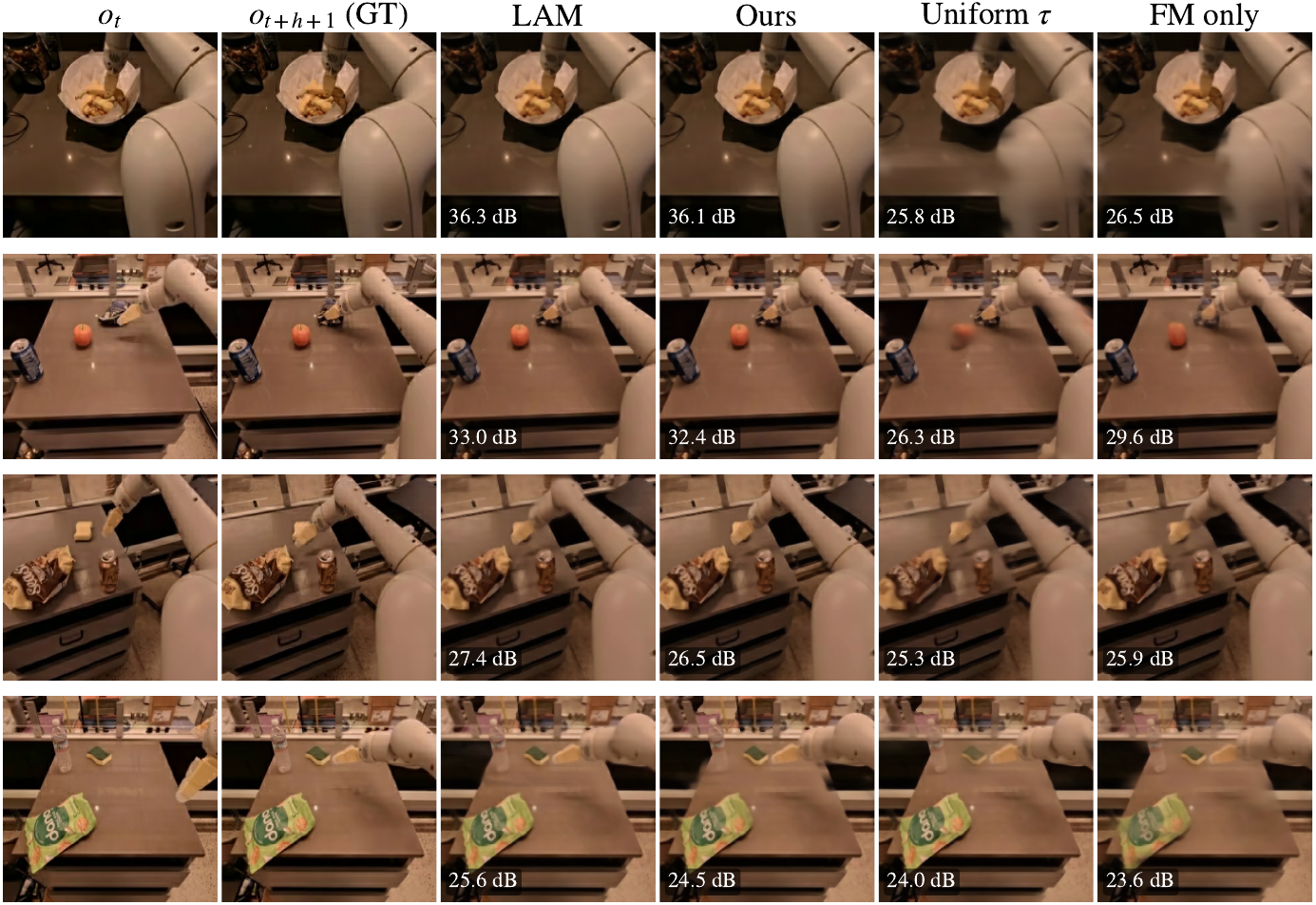}
    \caption{Additional latent reconstructions on Google Robot (Fractal). Columns follow \figref{fig:sup_recon_egodex}.}
    \label{fig:sup_recon_fractal}
\end{figure}

\figref{fig:z_pred_quality} shows one LAM-decoder reconstruction on EgoDex and WidowX. \figref{fig:sup_recon_egodex}-\ref{fig:sup_recon_fractal} provide four additional examples each on EgoDex, WidowX (BridgeData~V2), and Google Robot (Fractal). Across these examples, \ours{} reconstructs the future close to the LAM teacher in PSNR while the Uniform~$\tau$ and FM-only variants degrade, so the flow-adaptive objective preserves the teacher's predictive content across embodiments.

\end{document}